\documentclass[sigconf,nonacm]{acmart}

\usepackage{graphicx}
\usepackage{amsmath}
\usepackage{booktabs}
\usepackage{epigraph}
\usepackage{algorithm}
\usepackage{algpseudocode}
\usepackage{comment}
\usepackage{enumitem}
\setlist{nolistsep}
\usepackage{dblfloatfix}
\usepackage{subcaption}
\usepackage{array}
\usepackage{multirow}
\usepackage[most]{tcolorbox}
\tcbuselibrary{fitting}

\definecolor{columbiablue}{rgb}{0.61, 0.87, 1.0}
\definecolor{mossgreen}{rgb}{0.68, 0.87, 0.68}
\definecolor{darkgreen}{RGB}{0, 128, 0}
\definecolor{lightsalmonpink}{RGB}{255, 153, 153} 
\definecolor{gold}{RGB}{255, 215, 0} 
\definecolor{darkcolumbiablue}{rgb}{0.3, 0.55, 0.75}
\definecolor{darkgold}{rgb}{0.85, 0.67, 0.0}

\AtBeginDocument{%
  }

\begin{document}

\title{ProbTalk3D: Non-Deterministic Emotion Controllable Speech-Driven 3D Facial Animation Synthesis Using VQ-VAE}


\author{Sichun Wu}
\affiliation{%
  \institution{Utrecht University}
  \city{Utrecht}
  \country{The Netherlands}}
\email{wsc462@gmail.com}

\author{Kazi Injamamul Haque}
\affiliation{%
  \institution{Utrecht University}
  \city{Utrecht}
  \country{The Netherlands}}
\email{k.i.haque@uu.nl}

\author{Zerrin Yumak}
\affiliation{%
  \institution{Utrecht University}
  \city{Utrecht}
  \country{The Netherlands}}
\email{z.yumak@uu.nl}

\renewcommand{\shortauthors}{Wu et al.}

\begin{abstract}
Audio-driven 3D facial animation synthesis has been an active field of research with attention from both academia and industry. While there are promising results in this area, recent approaches largely focus on lip-sync and identity control, neglecting the role of emotions and emotion control in the generative process. That is mainly due to the lack of emotionally rich facial animation data and algorithms that can synthesize speech animations with emotional expressions at the same time. In addition, majority of the models are deterministic, meaning given the same audio input, they produce the same output motion. We argue that emotions and non-determinism are crucial to generate diverse and emotionally-rich facial animations. In this paper, we propose ProbTalk3D a non-deterministic neural network approach for emotion controllable speech-driven 3D facial animation synthesis using a two-stage VQ-VAE model and an emotionally rich facial animation dataset 3DMEAD. We provide an extensive comparative analysis of our model against the recent 3D facial animation synthesis approaches, by evaluating the results objectively, qualitatively, and with a perceptual user study. We highlight several objective metrics that are more suitable for evaluating stochastic outputs and use both in-the-wild and ground truth data for subjective evaluation. To our knowledge, that is the first non-deterministic 3D facial animation synthesis method incorporating a rich emotion dataset and emotion control with emotion labels and intensity levels. Our evaluation demonstrates that the proposed model achieves superior performance compared to state-of-the-art emotion-controlled, deterministic and non-deterministic models. We recommend watching the supplementary video for quality judgement. The entire codebase is publicly available\footnote[1]{\url{https://github.com/uuembodiedsocialai/ProbTalk3D/}}. 
\end{abstract}



\begin{teaserfigure}
\centering
  \includegraphics[width=0.8\textwidth]{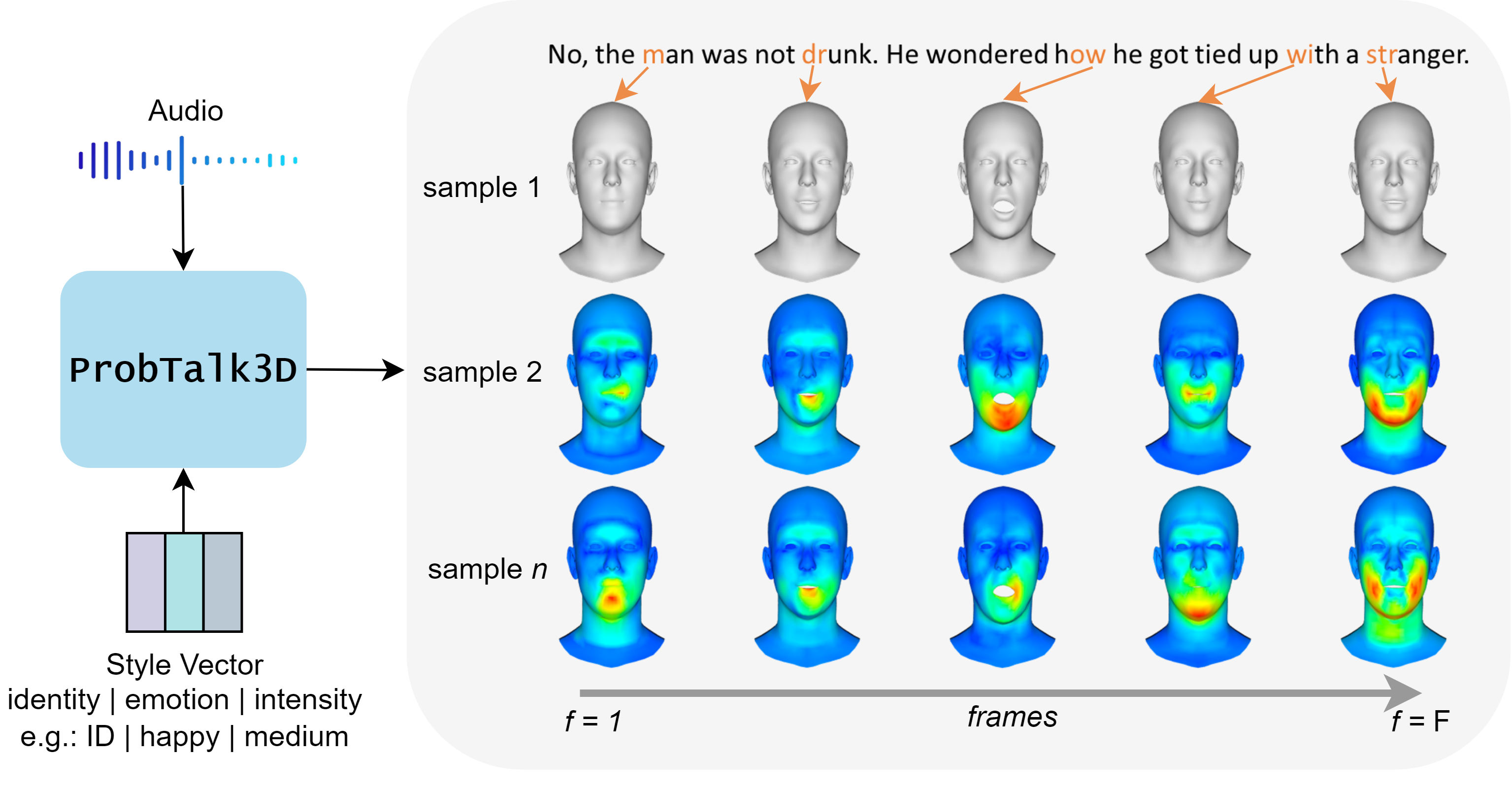}
  \caption{ProbTalk3D can non-deterministically synthesize 3D facial animation given audio, subject ID, emotion class, and emotion intensity as inputs. The generated samples using the same set of inputs of ProbTalk3D demonstrate diversity across multiple generations while ensuring proper lip-synchrony, emotional expressivity and visual quality.}
  \Description{Description of the teaser figure of proposed the proposed model, ProbTalk3D.}
  \label{fig:teaser}
\end{teaserfigure}

\maketitle

\section{Introduction}
3D facial animation is not only crucial in film-making and game production but also in a variety of XR applications that involve digital humans. Creating facial animations for 3D characters require lots of manual work from skilled technical artists or relies on performance capture pipelines. Researchers are actively addressing the challenges of 3D animation synthesis, aiming to minimize the manual effort. In particular, the relationship between speech and facial animation has been extensively studied and shown to be a promising direction \cite{karras2017audio, cudeiro2019voca, xing2023codetalker, FaceDiffuser_Stan_MIG2023}. However, recent speech-driven 3D facial animation synthesis methods mostly focus on lip-sync and identity control neglecting the role of emotions and emotion control. They are also mostly deterministic methods limiting the generation of diverse facial animations. These limitations are what we address in this paper.

Early 3D facial animation generation synthesis methods use procedural approaches \cite{edwards2016jali, Charalambous2019} by employing linguistic rules to map phonemes to visemes (visual counterparts of phonemes). Despite their artist-friendly features, these models require defining explicit rules and their output is limited to lip-sync only. In recent years, end-to-end deep learning methods have demonstrated their effectiveness in speech-driven 3D facial animation synthesis \cite{karras2017audio, taylor2017deep, cudeiro2019voca, fan2022faceformer, richard2021meshtalk,xing2023codetalker, thambiraja2023imitator, FaceXHuBERT_Haque_ICMI23} generalizing to diverse audio inputs and different languages. They rely on 3D vertex-based datasets such as VOCASET \cite{cudeiro2019voca}, Multiface \cite{wuu2022multiface} and BIWI \cite{fanelli20103biwi} with limited size and emotional variations. Lately non-deterministic approaches \cite{ng2022learning2listen, FaceDiffuser_Stan_MIG2023, yang2024probabilistic, zhao2024media2face, FaceTalk_Aneja_2024_CVPR, sun2023diffposetalk} have been proposed and there is an increasing number of papers that are listed on Arxiv \cite{ma2024diffspeaker, chen2023diffusiontalker, 3diface, park2023df}. A few approaches focus on rigged characters and blendshapes such as \cite{Voice2FaceEA, FaceDiffuser_Stan_MIG2023, park2023said}. Another group of work focus on holistic motion generation including face and body \cite{yi2023generating, emage, Diffsheg, ng2024audio2photoreal}. While there is an increasing number of papers in this area, only a few papers focus on emotion control and emotionally-rich animation generation \cite{karras2017audio, peng2023emotalk, emote, zhao2024media2face}. 

The work from Karras et al. \cite{karras2017audio} although focusing on emotions, is based on a small dataset with two actors and there is no explicit emotion control. The closest to our work are EmoTalk \cite{peng2023emotalk} and EMOTE  \cite{emote}. They both rely on 3D datasets constructed from 2D videos with EMOTE having superior visual quality and wider range of emotions. Similar to EMOTE, we opt for using the 3DMEAD dataset as it provides facial animations with various emotions at different intensity levels and for several identities. However, none of these methods are non-deterministic and they cannot generate diverse outputs given the same audio input. A recent method \cite{zhao2024media2face} introduces a 4D high-quality dataset with emotion variations, however this dataset and codebase is not publicly available yet.

\noindent The contributions of our work are enumerated below: 
\begin{itemize}[leftmargin=*]
    \item A novel two-stage probabilistic (i.e. non-deterministic) emotion controllable speech-driven 3D facial animation synthesis model based on VQ-VAE, producing diverse yet high-quality facial animations by learning a latent representation of emotional speech animation. 
    \item Extensive comparative analysis of our approach with respect to recent non-deterministic approaches using an enhanced list of objective metrics following \cite{yang2024probabilistic}.
    \item Qualitative evaluation and perceptual user study to demonstrate the superior perceptual quality of our model's results compared to the state-of-the-art emotion-control enabled as well as non-deterministic models.
\end{itemize}

\section{Related Work}
This section provides a review of related work on speech-driven 3D facial animation synthesis using deep learning algorithms. Although there is a vast amount of deep learning methods that are used to generate 2D facial animation \cite{Xinya2022eamm, Stypulkowski_2024_WACV, zhang2023sadtalker}, that is out of the scope of this paper. Another group of work focus on learning representations of facial animation, tracking and reconstruction \cite{FLAME, Giebenhain_2024_CVPR, EMOCA:CVPR:2021, Egger2020}. 3D speech-driven facial animation methods typically relied on phoneme-based procedural approaches \cite{JALI, Charalambous2019} or intermediary representations \cite{Taylor2012}. With the end-to-end deep learning approaches \cite{karras2017audio, taylor2017deep, cudeiro2019voca, fan2022faceformer, richard2021meshtalk,xing2023codetalker, thambiraja2023imitator, FaceXHuBERT_Haque_ICMI23}, a new era in 3D speech-driven facial animation started. Although providing promising results, these models are deterministic models limiting the generation of diverse outputs given the same speech input. They also do not provide explicit emotion control while most of them provide identity control. In this paper, we focus on non-deterministic speech-driven and emotion-controllable 3D facial animation synthesis. We will present the state-of-the-art in these two categories in the following two subsections.

\subsection{Non-Deterministic 3D Facial Animation}
In recent years, researchers have increasingly focused on the non-deterministic aspects of human motion both for body \cite{tevet2022human, alexanderson2023listen, Ao2023GestureDiffuCLIP, yi2023generating, Chhatre_2024_CVPR} and facial motion \cite{ng2022learning2listen, yang2024probabilistic, FaceDiffuser_Stan_MIG2023, zhao2024media2face, FaceTalk_Aneja_2024_CVPR, sun2023diffposetalk} as well as for holistic animation \cite{yi2023generating, emage, Diffsheg, ng2024audio2photoreal}. Non-deterministic models are mostly based on Variational Auto Encoders (VAEs), Vector Quantized Variational Auto Encoders (VQ-VAEs) or diffusion models. Compared to Generative Adversarial Networks (GANs), VAE-based and diffusion models stand out by generating diverse outputs by explicitly modeling the underlying data distribution \cite{Stypulkowski_2024_WACV}. Unlike traditional VAEs that encode input into a continuous space, VQ-VAEs represent data using discrete codebook embeddings preventing the posterior collapse problem \cite{VQVAE}. Richard et al. \cite{richard2021audio} propose a Temporal Convolutional Network (TCNN)-based VAE method to drive Codec Avatars \cite{lombardi2018deep}, while Voice2Face \cite{Voice2FaceEA} proposes an LSTM-based conditional VAE. MeshTalk \cite{richard2021meshtalk} learns a categorical latent space while CodeTalker \cite{xing2023codetalker} uses VQ-VAE for learning a discrete code space. However, these methods were not explicitly declared non-deterministic and do not provide an evaluation on the diverse outputs generated. Learning to Listen \cite{ng2022learning2listen} employs a transformer-based VQ-VAE for facial animation synthesis in dyadic conversations and introduces evaluation metrics to assess diversity including diverseness within and across sequences. Yang et al. \cite{yang2024probabilistic} uses Residual Vector-Quantized (RVQ) codebook achieving improved diversity and high fidelity in facial motion. They provide an extensive benchmarking framework and introduce novel evaluation metrics. FaceDiffuser \cite{FaceDiffuser_Stan_MIG2023} is the first model to apply diffusion models for the speech-driven 3D facial animation synthesis task. They introduce a new diversity metric that allows the comparison of this model to other deterministic models by measuring variation over identity instead of audio input. Facetalker \cite{aneja2023facetalk} employs a transformer-based diffusion model and predicts animations of neural parametric head models (NPHMs) \cite{giebenhain2023mononphm} offering detailed representations of the human head. They measure diversity using metrics similar to the ones used in the body animation domain \cite{ren2023diffusion}. DiffPoseTalk \cite{sun2023diffposetalk} introduces a new dataset that also includes head poses which is constructed from 2D videos and uses a combination of diffusion and transformer networks. EMAGE \cite{emage} being a holistic model encode different body parts using VQ-VAEs and uses various evaluation metrics including diversity based on \cite{Jing2021}. Another holistic model Audio2PhotoReal \cite{ng2024audio2photoreal} employs a diffusion model for the face. Although the above methods are non-deterministic, they do not provide explicit emotion control for speech-driven 3D facial animation.

\subsection{Emotion-Controllable 3D Facial Animation}
 There are a few papers in this space \cite{karras2017audio, peng2023emotalk, emote, zhao2024media2face}. Karras et al. \cite{karras2017audio} proposed an end-to-end method using CNNs aiming to resolve the ambiguity in mapping between audio and face by introducing an additional emotion component to the network. The dataset is based on two actors and cannot handle identity variations. In addition, there is no explicit emotion control but emotions are learned from data. The advantage is that the dataset is collected with a commercial high-end 4D performance capture system. EmoTalk \cite{peng2023emotalk} introduces an emotion disentangling encoder to disentangle the emotion and content in the speech using a cross-reconstruction loss. In contrast with Karras et al. \cite{karras2017audio}, they use a dataset that is semantically annotated with emotion labels. Given the emotion and content features, personal style and emotion control features, an emotion-guided multi-head attention decoder generates the output motion. Considering the limited availability of emotional 3D audio-visual datasets, EmoTalk addresses this gap by creating their own dataset 3D-ETF using two 2D datasets RAVDESS \cite{livingstone2018ryerson} and HDTF \cite{zhang2021flow}. They use the "Live Link Face" application to map input videos to blendshape parameters. They also include a blendshape to FLAME \cite{FLAME} parameters converter which allows to transfer facial expressions across different virtual characters. EMOTE \cite{emote} adopts a similar approach to EmoTalk and constructs a new dataset 3DMEAD based on the 2D dataset MEAD \cite{kaisiyuan2020mead} including annotated 8 basic emotion types, 3 emotion intensity types per emotion type, and speaker identities. To simplify the problem, they first learn a motion prior based on FLAME parameters changing in time using a temporal VAE. In the audio-driven training stage, they combine audio, identity and emotion features using transformer encoder and decoder structures to infer the motion. In contrast with FaceFormer \cite{fan2022faceformer} and CodeTalker \cite{xing2023codetalker}, they employ a non-autoregressive model for improved efficiency. Although EMOTE does not provide explicit objective and subjective evaluations with respect to EmoTalk, visual results indicates better visual quality. None of these models can produce diverse results in a non-deterministic manner. A recent paper Media2Face \cite{zhao2024media2face} proposes a two-stage model and a 4D dataset M2F-D. Different from EmoTalk and EMOTE, emotion control is not categorical emotions but rely on CLIP \cite{CLIP} text encoding. In the first stage, a latent space is learned using geometry and expression VAE models. The extracted latent codes are used to augment the dataset with 2D videos such as MEAD \cite{kaisiyuan2020mead}, RAVDESS \cite{livingstone2018ryerson} and HDTF \cite{zhang2021flow} similar to EmoTalk and EMOTE. In the second stage a transformer-based diffusion model is used. Although Media2Face is essentially a non-deterministic generative model, they do not employ an explicit diversity metric and use only FDD (Upper face Dynamics Deviation) metric similar to CodeTalker \cite{xing2023codetalker}. The visual quality of the results are good, however the dataset and codebase is not yet available for direct comparison.

\begin{figure*}[t]
  \centering
    \includegraphics[width=0.85\linewidth,clip]{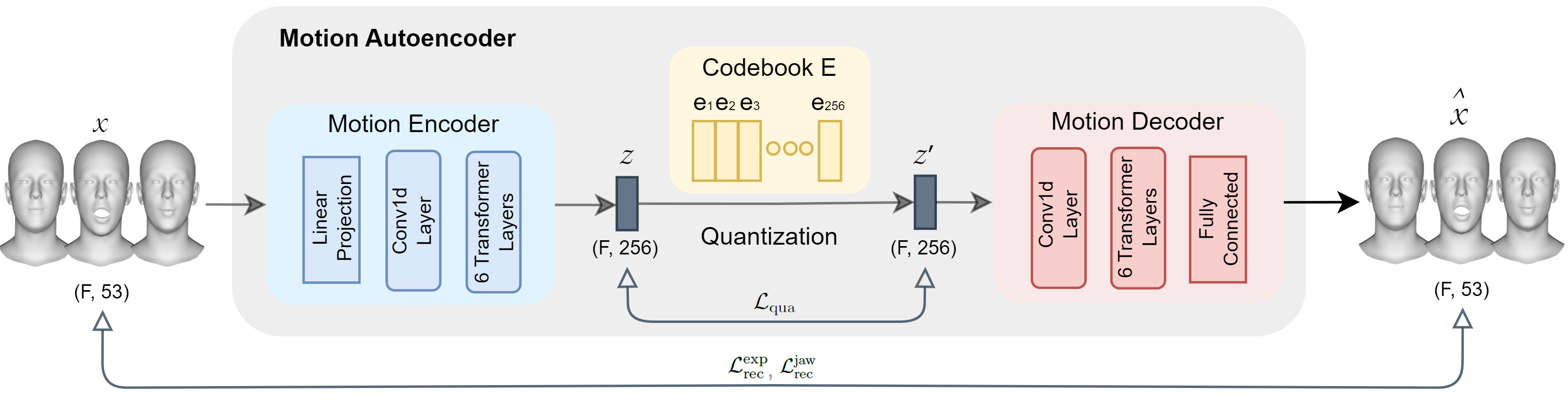}
    \caption{Stage 1: The motion autoencoder is trained by utilizing VQ-VAE to learn a motion prior in terms of discrete quantized codebook embeddings, $E$ . The autoencoder consists of a motion encoder that encodes the 53-dimensional temporal facial animation data into 256-dimensional latent space, $z$ that undergoes quantization to produce $z'$. With the help of a motion decoder, $z'$ gets decoded into facial animation data, $\hat{\mathcal{X}}$ with the same shape as the $\mathcal{X}$.} 
  \Description{ProbTalk3D stage 1 detailed figure.}
  \label{fig:training1}
\end{figure*}

\section{Methodology}
In this section, we describe our methodology including a description of the dataset, problem formulation, details about the proposed model, and two other non-deterministic approaches (VAE and diffusion-based) that are used to compare and evaluate our proposed model. The results and the comparative analysis are then presented in Sec. \ref{sec:results}.

\subsection{Dataset: 3DMEAD}
3DMEAD dataset is reconstructed from 2D audio-visual dataset MEAD \cite{kaisiyuan2020mead}. The 3D reconstruction from the 2D videos was carried out using DECA \cite{DECA:Siggraph2021} and MICA \cite{MICA} methods which was first introduced with the EMOTE \cite{emote} paper. 3DMEAD dataset includes 3D reconstructions of 47 subjects speaking in English, comprising eight emotions at three intensity levels. The emotion categories include neutral, happy, sad, surprised, fear, disgusted, angry, and contempt. Except for the neutral class, each emotion category has three intensity levels: weak, medium, and strong. Every subject contributes 30 short sentences for seven basic emotions, each expressed at three aforementioned intensity levels, along with an additional 40 sentences with the neutral emotion.

We choose 3DMEAD for our experiments as it offers a relatively large-scale, high-quality facial animation data with coverage of diverse emotions. Motion data is sampled at $25 fps$. Each frame in the dataset is represented using FLAME \cite{li2017flame} 3D Morphable Model (3DMM) parameters $\{\beta, \theta, \psi \} \in \mathbb{R}^{406}$, where $\beta \in \mathbb{R}^{300}$ denotes the face shape, $\theta_{jaw} \in \mathbb{R}^{3}$ denotes the Euler angle rotation (x,y,z) of the jaw bone, $\theta_{global} \in \mathbb{R}^{3}$ denotes the global head pose, and $\psi \in \mathbb{R}^{100}$ denotes the expression parameters. However, similar to EMOTE, we utilize only $\{\psi, \theta_{jaw}\} \in \mathbb{R}^{53}$ for model training, where $\psi \in \mathbb{R}^{50}$ representing the first 50 of the total 100 expression parameters. The original training configuration of EMOTE splits the dataset into training-validation-test sets keeping all sequences per subject while having different subjects in each set. In this way, there is no ground truth data to perform a quantitative evaluation for which EMOTE only conducted a perceptual user study. In contrast, our proposed split keeps a small number of sequences from each training subject for validation and testing allowing comparison of generated samples with respect to the ground truth. Although our training is done on fewer sequences than EMOTE, given the large scale of 3DMEAD dataset, we demonstrate that this split provides sufficient information for effective training and generation of animations perceptually superior in comparison to the EMOTE model. More details about the dataset split can be found in the supplementary material.

\subsection{Problem Formulation}
The task is to generate facial animation sequences based on audio and style inputs. To this end, we propose a supervised neural network model training approach to learn from data so that after training, we can predict the facial motion on any arbitrary unseen inputs. To train such a model, we leverage the audio-motion pairs in 3DMEAD dataset. The problem can be formulated as follows-

Let $\mathcal{X} = \{\mathcal{X}^f\}_{f=1}^F$ represent a facial animation sequence containing $F$ frames, paired with audio sequence $a$. Each sequence is also annotated in terms of style, $\mathcal{C}$. Hence, the training process to learn the weights of the defined model, $M$ can be expressed as - 
\begin{equation} \label{train_equation}
\hat{\mathcal{X}} = M (\mathcal{X}, a, \mathcal{C})
\end{equation}
where given audio $a$ and style $\mathcal{C}$, the model weights of $M$ are optimized to be able to predict $\hat{\mathcal{X}}$ that resembles the real facial animation data, $\mathcal{X}$.

After training, the inference process on arbitrary audio and style inputs can be formulated as- 
\begin{equation} \label{inference_equation}
\hat{\mathcal{X}} = M (a, \mathcal{C})
\end{equation}

The facial motion is defined as, $\mathcal{X}$ or $\hat{\mathcal{X}} \in \mathbb{R}^{F \times P}$, where $F$ represents the number of total visual frames and $P$ represents the dimension of animation data. While training the model with 3DMEAD, $P = 53$ comprises the first 50 FLAME expression parameters and 3 jaw parameters (x,y,z Euler rotation of the jaw bone). The training process is conditioned on a style vector $\mathcal{C} = [c_{id} | c_{emo} | c_{int}]$, where $c_{id}, c_{emo}, c_{int}$ are concatenated one-hot vectors representing subject identity, emotion class, and intensity category respectively.

\begin{figure*}
  \centering
    \includegraphics[width=0.87\linewidth,clip]{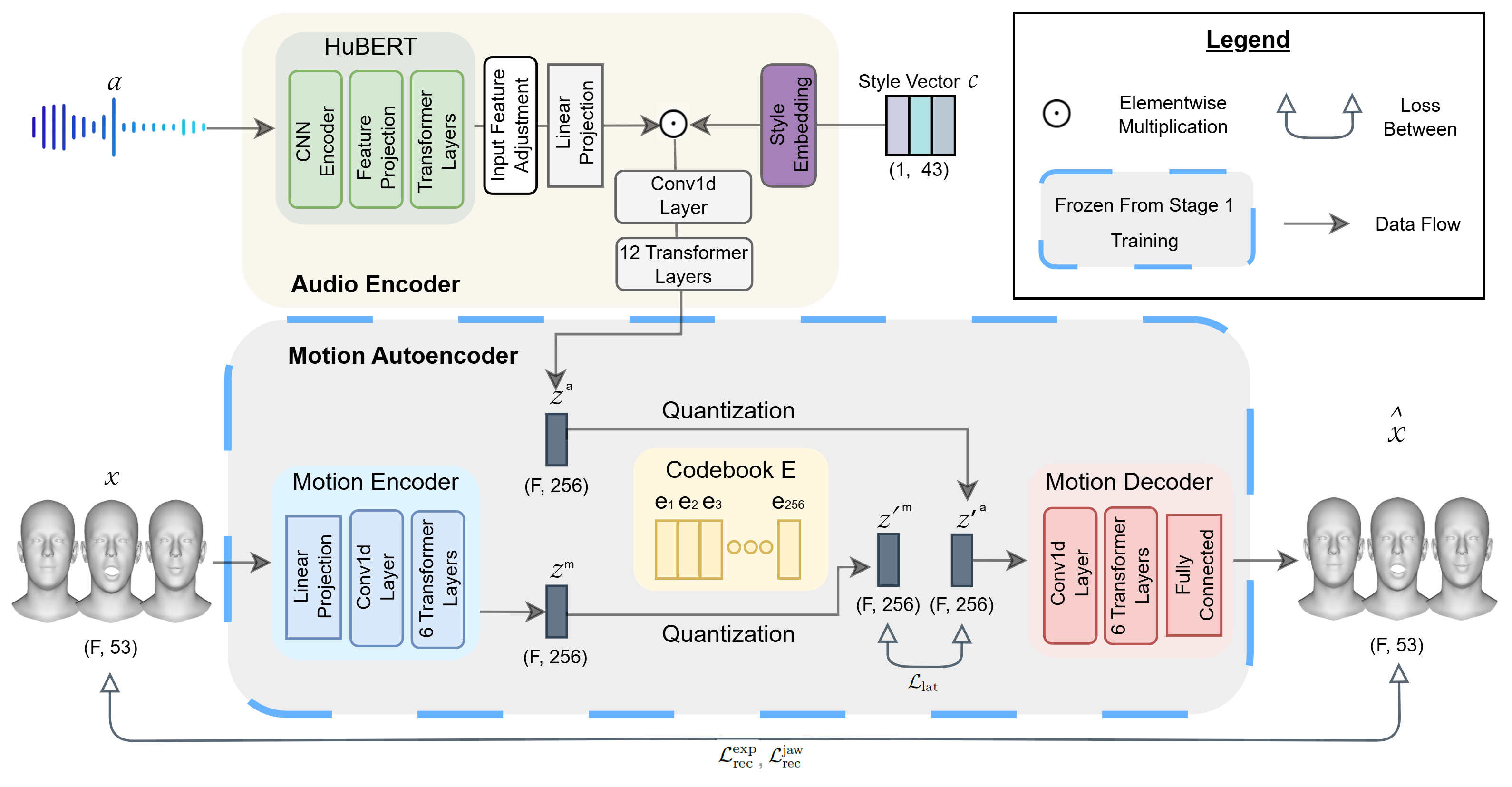}
    \caption{Stage 2: We keep the motion autoencoder trained in the previous stage frozen and train the HuBERT based Audio Encoder in such a way that given audio $a$, it produces the quantized audio latent, $z'^{\text{a}}$ from audio encoder output $z^{\text{a}}$ such that $z'^{\text{a}}$ closely resembles the quantized motion latent, $z'^{\text{m}}$. The Style Vector, $\mathcal{C}$ (i.e. concatenated one-hot vectors containing information about subject ID, emotion class and emotion intensity) is used to learn the Style Embedding that is fused with the encoded audio information. Unlike stage 1 training, the audio latent $z^{\text{a}}$ is used to get the quantized codebook latent $z'^{\text{a}}$ that is then decoded into facial animation $\hat{\mathcal{X}}$, utilizing the frozen motion decoder.} 
  \label{fig:training2}
\end{figure*}

\subsection{Proposed Model: ProbTalk3D}
Our proposed model ProbTalk3D follows a 2-stage training process similar to CodeTalker \cite{xing2023codetalker} and EMOTE \cite{emote} where in the first stage, we learn a motion prior with a motion autoencoder and in the second stage, we train a speech and style conditioned network by leveraging the pretrained HuBERT audio encoder\cite{hubert} and the motion prior from stage 1. Different from EMOTE \cite{emote}, our proposed model produces diverse outputs non-deterministically with less training data, a less complex and more efficient architecture. The perceptual losses from EMOTE (lip-reading and video-emotion loss) require the ground truth performance videos from the original MEAD dataset, and these involve extra processing that ours does not require. Instead we use quantization and reconstruction losses in the first and second stage.

\subsubsection{Stage 1: Motion Autoencoder}
The Motion Autoencoder consists of a motion encoder and a decoder aimed to learn a facial motion prior, leveraging the concept of Vector Quantized Variational Autoencoder (VQ-VAE) as laid out in Fig. \ref{fig:training1}. The Motion Encoder maps the motion input into a latent space during training. We employ transformers, which are proven to effectively capture and learn temporal context. Specifically, the encoder contains a linear projection layer, a 1D convolutional layer, and 6 transformer layers with residual attention and positional encoding. Given an input $X \in \mathbb{R}^{F \times P}$, the Motion Encoder encodes the data to a latent vector $z \in \mathbb{R}^{F \times 256}$. The encoded motion, $z$ undergoes vector quantization and learns a discrete latent embedding codebook, $E$. We configure $E$ to have 256 latent embeddings with each embedding dimension being 128. This setup implies a codebook size of 256, indicating that the motion is categorized into 256 types and each category is represented by a 128-dimensional vector. From the codebook, we find the embedding that is close to $z$ in terms of distance in the quantization process. This selected embedding is then reshaped to align with the dimension of $z$, resulting in a quantized latent motion feature, $z'$. More details on the background knowledge of VQ-VAE can be found in the supplementary material. 
The Motion Decoder consists of a 1D convolutional layer followed by 6 transformer layers. Additionally, a fully connected layer projects the hidden units back to the original input dimension. The Motion Decoder accepts $z' \in \mathbb{R}^{N \times 256}$ as input and processes it to yield $\hat{\mathcal{X}} \in \mathbb{R}^{F \times P}$. 
The architecture shares similarities with recent VQ-VAE based works \cite{ng2022learning2listen, xing2023codetalker}. However, unlike these models that autoregressively predict future frames based on previously generated frames for a given sequence, ours is non-autoregressive similar to EMOTE \cite{emote}, notably improving training and inference efficiency.

\paragraph{Loss function} We define the loss function for training stage 1 in Eq.\eqref{equation_vqvae_mloss_stage1}. There are three weighted loss terms- (i) $\mathcal{L}_{qua}$, that represents the standard quantization loss with codebook loss term and commitment loss term proposed for VQ-VAE models (see Eq.\eqref{equation_loss_qua}), (ii) $\mathcal{L}_{rec}^{exp}$, that computes the $\mathcal{L}_1$ loss between decoded motion and ground truth in terms of the 50 expression parameters, (iii) $\mathcal{L}_{rec}^{jaw}$, that computes the $\mathcal{L}_1$ loss between decoded motion and ground truth in terms of the 3 jaw parameters.

\begin{equation} \label{equation_loss_qua}
\mathcal{L_\text{qua}} = \|\text{sg}[z] - {z'}\|_2^2 + \beta \|z - \text{sg}[z']\|_2^2
\end{equation}

\begin{equation}  \label{equation_vqvae_mloss_stage1}
\mathcal{L_\text{stage1}} = \lambda_\text{qua}\mathcal{L}_\text{qua} + \lambda_\text{rec}^\text{exp}\mathcal{L}_\text{rec}^\text{exp}  +  \lambda_\text{rec}^\text{jaw}\mathcal{L}_\text{rec}^\text{jaw}  
\end{equation}

The weights of the loss function in Eq.\eqref{equation_vqvae_mloss_stage1} are empirically set as follows: $\lambda_\text{qua} = 1.5$, $\lambda_\text{rec}^\text{exp}=0.5$, and $\lambda_\text{rec}^\text{jaw}=0.1$.

\subsubsection{Stage 2: Speech and Emotion Conditioned}
After learning the motion prior in stage 1, speech and emotion conditioned stage 2 is trained which consists of an audio encoder that encodes the continuous raw audio/speech data into discrete hidden representations and fuses the speaking style (i.e. style embedding obtained using Style Vector, $\mathcal{C}$) with the encoded audio information. As shown in Fig.\ref{fig:training2}, the Motion Autoencoder trained in stage 1 is kept frozen in this stage to train the Audio Encoder in a supervised manner so that the latent representation, $z^{\text{a}}$ from the audio encoder produces a quantized latent $z'^{\text{a}}$ that closely resembles the quantized latent motion representation, $z'^{\text{m}}$.

\begin{figure*}
  \centering
    \includegraphics[width=0.9\linewidth,clip]{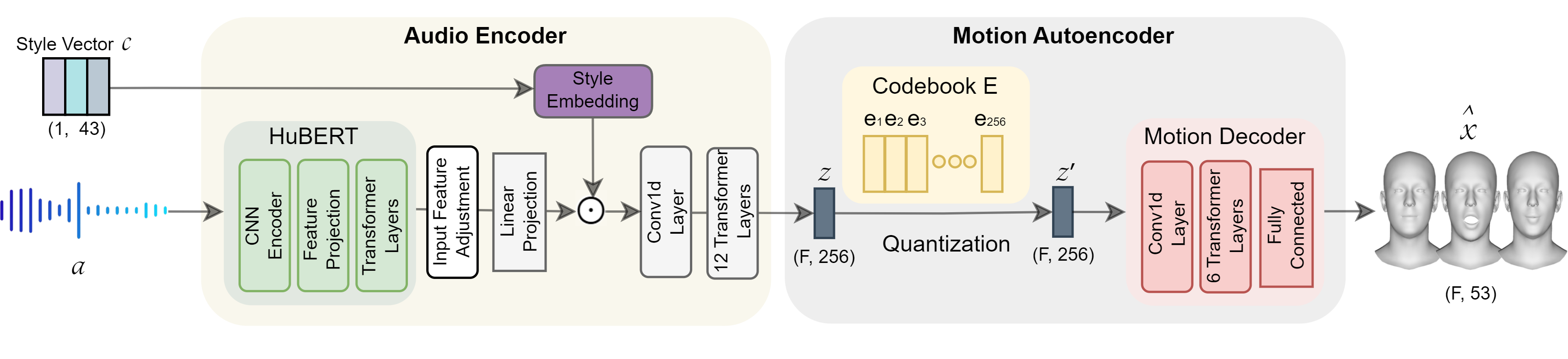}
    \caption{Inference: Given audio $a$ and Style Vector $\mathcal{C}$ (i.e. subject ID, emotion, intensity) as inputs, the audio encoder produces hidden representation $z$ that undergoes the quantization process based on the learned codebook embeddings $E$ and probabilistically produces $z'$. The motion decoder then decodes $z'$ into facial animation $\hat{\mathcal{X}}$.}  
  \Description{ProbTalk3D inference figure.}
  \label{fig:inference}
\end{figure*}

We employ pretrained HuBERT \cite{hubert}, a transformer based speech recognition model in our audio encoder, and made appropriate modifications to apply it to our downstream task of facial animation synthesis following \cite{FaceXHuBERT_Haque_ICMI23, FaceDiffuser_Stan_MIG2023}. After the last hidden layer of HuBERT, inspired by the aforementioned previous works, we adjust the input representation so it temporally aligns with the paired animation data and add a linear projection layer that projects the encoded hidden state into a 256-dimensional hidden state to be element-wisely multiplied with the Style Embedding. Style Embedding is obtained using a linear projection of concatenated one-hot vectors of the style annotations (subject ID, emotion, intensity). In our case, the style vector, $\mathcal{C}$ is a 43-dimensional vector (as we have 32 subjects, 8 basic emotions, and 3 emotion intensity categories) that is linearly transformed into a 256-dimensional Style Embedding to facilitate the fusion of audio and style information. After fusing the audio information and style information, we introduce a 1D convolutional layer followed by 12 transformer layers to get the latent representation $z^{\text{a}}$ that goes through the prior codebook quantization and the motion decoder to obtain the predicted motion, $\hat{\mathcal{X}}$.

\paragraph{Loss function}
The goal in stage 2 training is to get audio latent $z^{\text{a}}$ such that after quantization it produces $z'^{\text{a}}$ that closely resembles the quantized motion latent, $z'^{\text{m}}$. We construct $\mathcal{L}_{lat} = \mathcal{L}_1(z'^{\text{m}},  z'^{\text{a}})$ that computes the $\mathcal{L}_1$ loss between these quantized latent representations. Additionally, we further add the reconstruction losses. The overall loss function is then defined in Eq.\eqref{equation_vae_mloss_stage2}. The loss weights set for training are $\lambda_\text{lat}=1$, $\lambda_\text{rec}^\text{exp}=0.15$, and $\lambda_\text{rec}^\text{jaw}=0.1$.
\begin{equation}  \label{equation_vae_mloss_stage2}
\mathcal{L_\text{stage2}} = \lambda_\text{lat}\mathcal{L}_\text{lat} + \lambda_\text{rec}^\text{exp}\mathcal{L}_\text{rec}^\text{exp}  +  \lambda_\text{rec}^\text{jaw}\mathcal{L}_\text{rec}^\text{jaw}  
\end{equation}

\subsubsection{Inference}
During inference on unseen audio sequences, the \textit{Motion Encoder} in the Motion Autoencoder is not used as there is no ground truth motion to encode. The trained Audio Encoder in our model maps input audio to the latent representation $z$. Subsequently, the quantization process extracts an embedding from the learned codebook $E$, yielding $z'$, which is then passed to the decoder of the Motion Autoencoder to synthesize facial animation, $\hat{\mathcal{X}}$. Style $\mathcal{C}$ can be specified to generate a particular subject's speaking style with specific emotion class and emotion intensity. The inference process is illustrated in Fig.\ref{fig:inference}. During training iterations, our model is optimized to retrieve the closest learned codebook embedding index. We introduce non-deterministic output generation by incorporating a probabilistic sampling process for the codebook embedding index retrieval in the quantization step.

\subsubsection{Training Details}
Our proposed model is implemented using the PyTorch Lightning \cite{PT_lightning} framework and trained on a single NVIDIA A16 GPU. During stage-1 training of the Motion Autoencoder, we use the AdamW optimizer while the Adam optimizer is used in stage 2. We empirically choose the learning rate to be $1e^{-4}$ and $1e^{-5}$ for stage 1 and stage 2 respectively while we monitor the validation loss by employing an early stopping technique to mitigate overfitting in both stages. More specifically, if the validation loss does not improve in the next 5 epochs, we stop the training and save the model weights from the final epoch. For our proposed model, stage 1 converges after training for 21 epochs ($\approx 6.5$ hours) while stage 2 after 48 epochs ($\approx 42$ hours).  

\subsection{Comparison Model: VAE Based}
In order to compare our model with a VAE-based model, we trained a variant of ProbTalk3D by replacing the VQ-VAE structure with a VAE keeping the rest of the architecture the same, ensuring a fair comparison between VAE and VQ-VAE based models. Unlike VQ-VAE which learns a discrete codebook embedding in stage 1 training, VAE learns the mean, $\mu$ and covariance matrix $\Sigma$ of the latent space so that it follows a normal distribution $\phi = \mathcal{N}(\mu, \Sigma)$. 

Similar to ProbTalk3D, this model learns the motion prior and in stage 2 the learned mean $\mu$ and covariance matrix $\Sigma$ are used and reparameterized for decoding audio and style conditioned facial animation. We use a similar loss function to our proposed model, replacing only the quantization loss term, $\mathcal{L}_{qua}$ in Eq.\eqref{equation_vqvae_mloss_stage1} with a Kullback-Leibler (KL) Divergence loss, $\mathcal{L}_{KL}$ with $\lambda_{KL} = 1e^{-4}$, $\lambda_{rec}^{exp} = 1.5$, $\lambda_{rec}^{jaw} = 1$. 

\begin{table*}[t]
    \centering
    \begin{tabular}{ccccccc}
    \toprule
        Model & MVE$\downarrow$ & LVE $\downarrow$ & FDD$\downarrow$ & MEE $\downarrow$ & CE $\downarrow$  & Diversity$\uparrow$\\
         &  x$10^{-3} mm$ & x$10^{-4} mm$ & x$10^{-5} mm$  & x$10^{-4} mm$ &  x$10^{-4} mm$ & x$10^{-3} mm$\\
    \midrule
        FaceFormer\cite{fan2022faceformer} & 2.8548  & 2.0266 & 0.0659 & N/A & N/A & N/A\\
        CodeTalker\cite{xing2023codetalker} &  1.7121 & 1.5978 & 0.2036 & N/A & N/A & N/A\\
        CodeTalker-ND &  3.4783 & 3.1325 & 0.3089 & 3.1593 & 3.0205 & 0.1696\\
        FaceDiffuser - DDPM \cite{FaceDiffuser_Stan_MIG2023} & 1.3242 & 0.8938 & 0.0906 & 0.8848 & 0.8784  & 0.0449 \\
        FaceDiffuser - DDIM & 0.9334 & \textbf{0.4440} & 0.0625 & \textbf{0.4440} & \textbf{0.4438} & 0.0009 \\
        VAE-based & \textbf{0.6959}  & 0.5673 & \textbf{0.0010} & 0.5657 & 0.5550 & 0.0722\\
        ProbTalk3D (proposed) & 0.7243 & 0.6040 & 0.0415 & 0.5549 & 0.5227 & \textbf{0.3274}\\
    \midrule    
    \multicolumn{7}{c}{Ablation Study}  \\
    \midrule    
        VAE-based (w/o emo) & 1.3357  & 1.4576 & 0.1022 & 1.4529 & 1.4110 & 0.1158\\
        ProbTalk3D (w/o emo) & \textbf{0.7799} & \textbf{0.5794} & \textbf{0.0579} & \textbf{0.5323} & \textbf{0.4989} & \textbf{0.3136}\\
    \bottomrule
    \end{tabular}
    \caption{Quantitative evaluation results comparing our proposed model ProbTalk3D with VAE-based variant and a modified version of FaceDiffuser together with other SOTA models. Our proposed model achieves significantly better results regarding the Diversity metric while providing comparable results to other non-deterministic models with respect to other metrics. Non-deterministic models including ours (ProbTalk3D and VAE-based model) are better than deterministic models overall. The ablation study results related to Sec.\ref{sec:ablation_study} are also shown here.}
    \label{tab:quant_result}
\end{table*}

\subsection{Comparison Model: Diffusion Based}
For our comparative analysis, we chose state-of-the-art diffusion based method FaceDiffuser\cite{FaceDiffuser_Stan_MIG2023}. FaceDiffuser also uses HuBERT as an audio encoder which ensures similarity with respect to our model. In addition to the original model, we train FaceDiffuser on 3DMEAD with a modification of the original model structure in order to incorporate the emotion and intensity categories. We use 3D vertex coordinates instead of FLAME parameters for training FaceDiffuser as our experiments indicate that using low dimensional FLAME parameters data does not yield realistic facial animations for this architecture. To obtain the vertex coordinate data, scripts provided by FLAME\cite{li2017flame} are utilized to convert temporal FLAME parameters into temporal 3D vertex coordinates. In the revised FaceDiffuser model, the style embedding is multiplied earlier in the network with audio features extracted by HuBERT, rather than fusing it later in the network, before the last fully connected layer, as it was done in the original model. The style embedding includes information about subject identity, emotion class, and emotion intensity, using the same format as the style vector $C$ in our proposed model, while the original FaceDiffuser only uses one-hot vectors for subject identities. Our experiments show that integrating the style at an earlier stage generates better results. We define two versions: FaceDiffuser - DDPM and FaceDiffuser - DDIM (See Tab.\ref{tab:quant_result}). The former is trained with Denoising Diffusion Probabilistic Model (DDPM) \cite{ddpm} same as the original FaceDiffuser. We further improve the diffusion sampling efficiency of the modified model by changing the original one with Denoising Diffusion Implicit Model (DDIM) \cite{ho2020denoising}. Aside from these changes, the model structure and hyperparameters are identical to the V-FaceDiffuser in \cite{FaceDiffuser_Stan_MIG2023}.

\section{Results}
\label{sec:results}
In this section, we present the results of our proposed model and evaluate the model quantitatively, qualitatively, and with a perceptual user study. In addition to the relevant objective metrics from the literature \cite{xing2023codetalker, FaceDiffuser_Stan_MIG2023}, we use an extensive list of objective metrics following \cite{yang2024probabilistic}. The non-deterministic generation ability of our model is illustrated using a diversity metric, in line with recent probabilistic models \cite{yang2024probabilistic, ren2023diffusion}. Furthermore, qualitative evaluations provide visual demonstrations of animation quality and emotion control aspects. Following that, results of the user study are provided, aiming to evaluate the perceived realism, lip-synchrony, and emotional expressivity of the synthesized animations. Finally, the ablation study showcases the impact of incorporating emotion control into our model.

\subsection{Quantitative Evaluation}
We quantitatively evaluate our model performance based on multiple metrics. Mean Vertex Error (MVE), Lip Vertex Error (LVE) and Upper Face Dynamics Deviation (FDD) are calculated based on one sample output which are metrics commonly used for evaluating deterministic models. Mean Estimate Error (MEE), CE (Coverage Error) and Diversity are calculated over multiple samples providing a better picture for evaluating non-deterministic models. The widely applied evaluation metrics used by recent models mostly operate in vertex coordinate space. To compare accuracy with these models, we employ scripts provided by FLAME \cite{li2017flame} to convert predicted motion parameters into vertex coordinate space. In addition to VAE-based and Diffusion-based models, we compare our model to deterministic models FaceFormer\cite{fan2022faceformer} and CodeTalker\cite{xing2023codetalker} trained on 3DMEAD. To ensure a fair comparison, we also implement a non-deterministic version of CodeTalker called CodeTalker-ND by incorporating probabilistic sampling during inference. Retraining EMOTE from scratch on our slightly modified dataset split was not an option as we found the publicly available training codebase to be difficult to reproduce within a reasonable time frame, as also reported in \cite{zhao2024media2face}. Hence, we only provide subjective comparisons to EMOTE as explained in Sec.\ref{sec:qualitative}. The evaluation metrics are concisely described below.

\paragraph{\textbf{MVE}} 
Mean Vertex Error calculates the average Euclidean distance between predicted frames and ground truth across the entire test set. Let $N$ represent the total number of frames generated for all the test-set audio inputs. We denote $x_i$ as the ground truth of $i$-th frame, and $\hat{x}_i$ as the predicted frame. The computation of MVE can be expressed as follows:
\begin{equation} \label{equation_eva_mve}
\text{MVE} = \frac{1}{N} \sum_{i=1}^{N} \|x_i -\hat{x}_i \|
\end{equation}

\paragraph{\textbf{LVE}} 
Lip Vertex Error calculates the maximal $\mathcal{L}_2$ error between vertices belonging to the lip/mouth region of a predicted frame compared to the ground truth, and then computes the mean across all generated frames. As the facial topology is same for 3DMEAD and VOCASET, we adopt the same lip mask utilized in \cite{xing2023codetalker,FaceDiffuser_Stan_MIG2023}. For $N$ predicted frames from all test audio, let $x_{lip}^i$ denote the lip region vertices of a ground truth frame, and $\hat{x}_{lip}^i$ represent the same region of a predicted frame. Then LVE is computed by:
\begin{equation} \label{equation_eva_lve}
\text{LVE} = \frac{1}{N} \sum_{i=1}^{N} \max \|x_{lip}^i -\hat{x}_{lip}^i \|_2 
\end{equation}

\paragraph{\textbf{FDD}}
Upper Face Dynamic Deviation was proposed in \cite{xing2023codetalker}, it measures the variation of facial dynamics for motion sequences in comparison with ground truth. It gives an indication of how close the standard deviation (or upper face motion variation) of generated sequences (of test-set audios) is compared to the variation observed in ground truth.  

\paragraph{\textbf{MEE}}
Mean Estimate Error \cite{yang2024probabilistic} is proposed to assess how close the mean of the sampling distribution is to the ground truth. To compute this, we generate a set of samples $S = \{\hat{x_1}, \hat{x_2},...,\hat{x_{10}}\}$. For each test audio, we generate $10$ motion sequences and calculate the mean, $E(\hat{x})$ of these 10 samples. Then we compute the MEE using Eq.\eqref{equation_eva_mee} for all the test sequences and take the average across the number of test sequences. A smaller MEE indicates that the model is more effective at generating ground truth lip movements. This metric is more suitable for probabilistic and non-deterministic models because it considers a set of samples instead of one sample.
\begin{equation} \label{equation_eva_mee}
\text{MEE} = \text{LVE} (x, E(\hat{x}))
\end{equation}

\paragraph{\textbf{CE}}
Coverage Error measures how close the sampling distribution of a probabilistic model is to the ground truth \cite{yang2024probabilistic}.  In order to calculate CE for one test sequence, we generate $S$ containing $10$ samples similar to MEE using our model and take the minimum of LVE calculation between ground truth and generated samples using Eq.\eqref{equation_eva_ce}. Taking the mean of CE over all test sequences yields the final CE in Tab.\ref{tab:quant_result}. A probabilistic model with a smaller CE has predictions that cover the ground truth samples in terms of lip motion.

\begin{equation} \label{equation_eva_ce}
\text{CE} = \underset{\hat{x}\in S}{\min} \ \text{LVE}(x, \hat{x})
\end{equation} 

\begin{figure}[t]
\centering
\resizebox{\linewidth}{!}{
\begin{tabular}{ccccccc}
 & \textcolor{orange}{p}rice & \textcolor{orange}{str}anger & \textcolor{orange}{m}an & \textcolor{orange}{w}as &
\textcolor{orange}{dr}unk & \textcolor{orange}{how} \\
\raisebox{5\height}{GT} &
\includegraphics[scale=0.18]{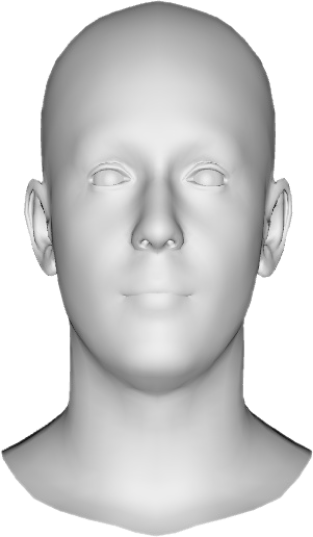} &
\includegraphics[scale=0.18]{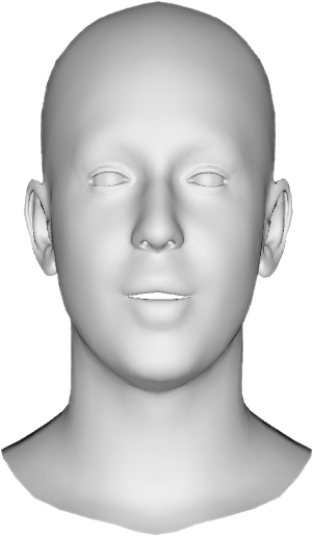} &
\includegraphics[scale=0.18]{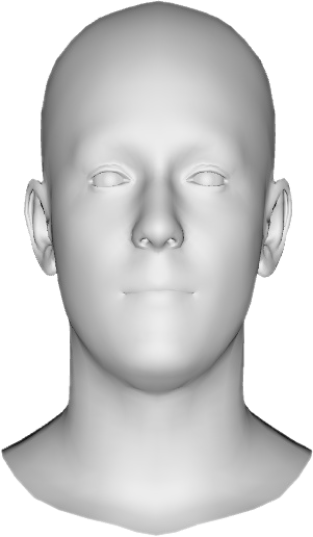} &
\includegraphics[scale=0.18]{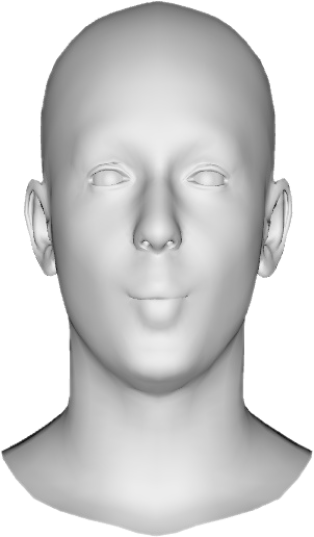} &
\includegraphics[scale=0.18]{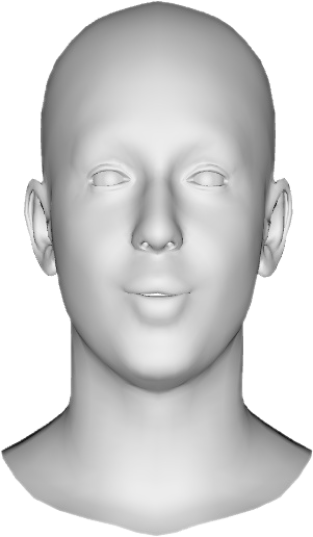} &
\includegraphics[scale=0.18]{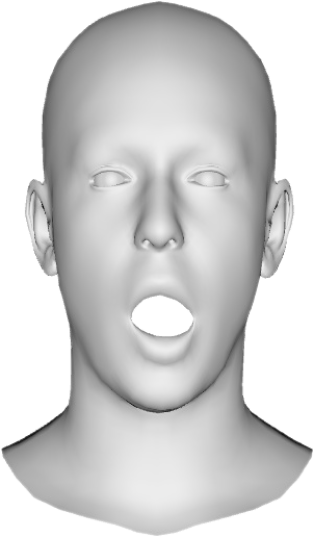} \\
\raisebox{5\height} {VAE based} &
\includegraphics[scale=0.18]{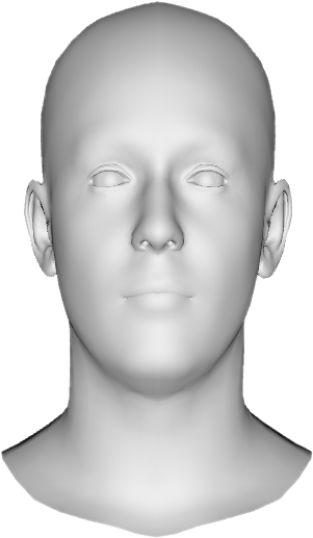} &
\includegraphics[scale=0.18]{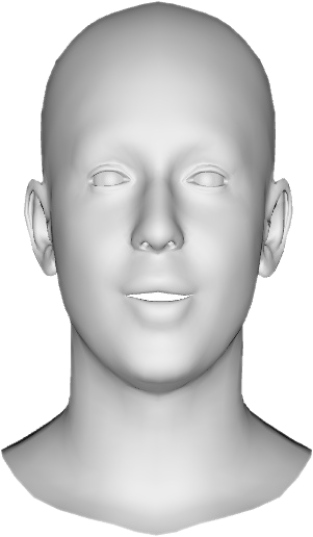} &
\includegraphics[scale=0.18]{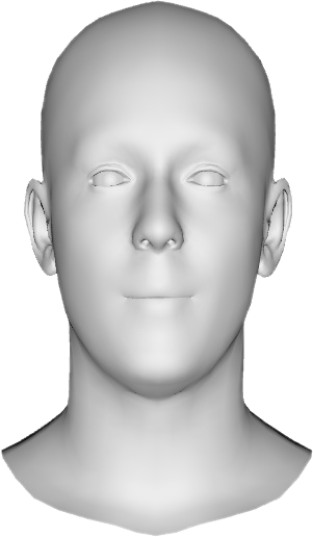} &
\includegraphics[scale=0.18]{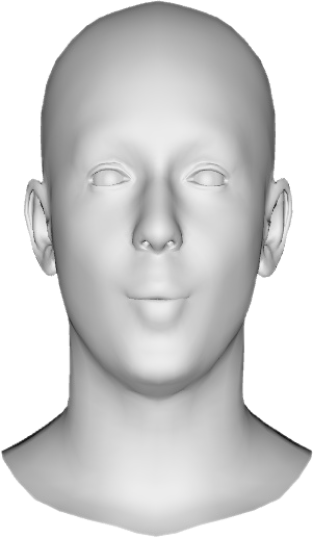} &
\includegraphics[scale=0.18]{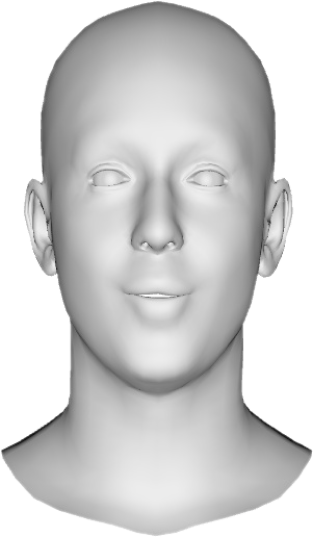} &
\includegraphics[scale=0.18]{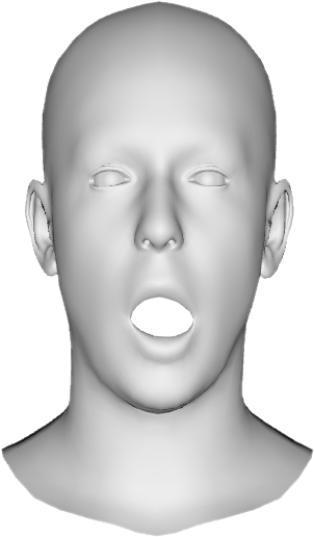} \\
\raisebox{5\height}{ FaceDiffuser} &
\includegraphics[scale=0.18]{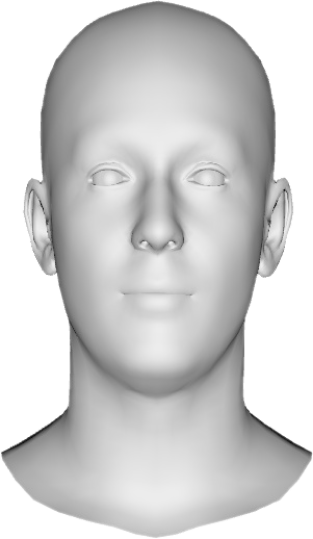} &
\includegraphics[scale=0.18]{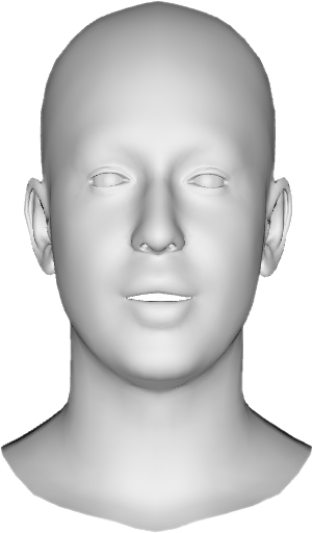} &
\includegraphics[scale=0.18]{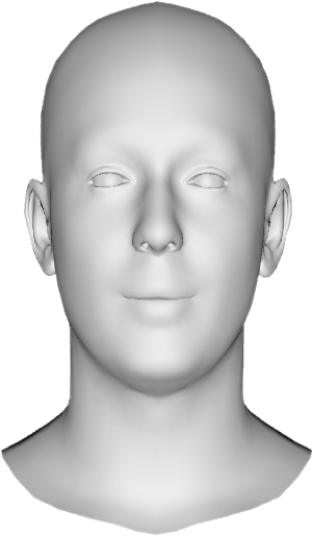} &
\includegraphics[scale=0.18]{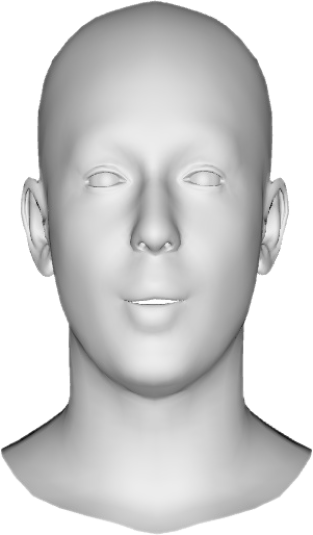} &
\includegraphics[scale=0.18]{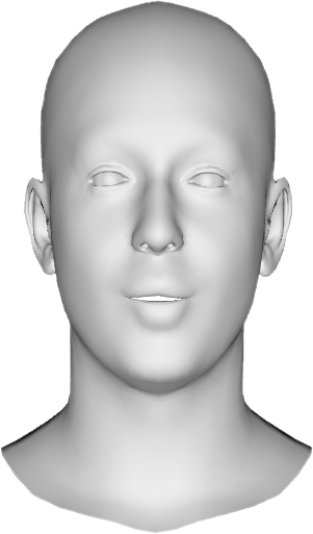} &
\includegraphics[scale=0.18]{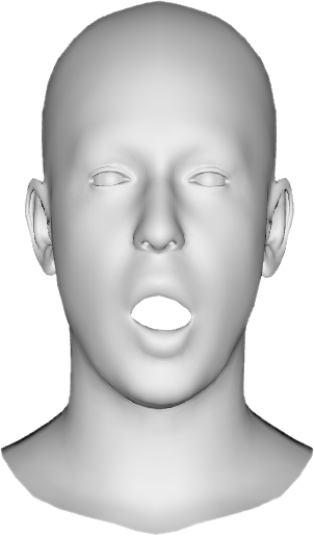} \\
\raisebox{5\height}{ EMOTE} &
\includegraphics[scale=0.18]{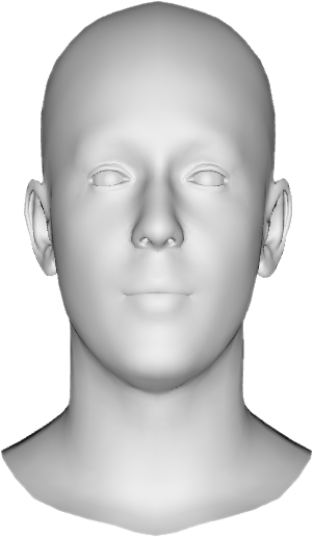} &
\includegraphics[scale=0.18]{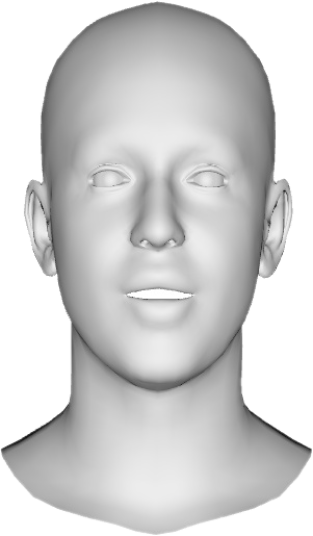} &
\includegraphics[scale=0.18]{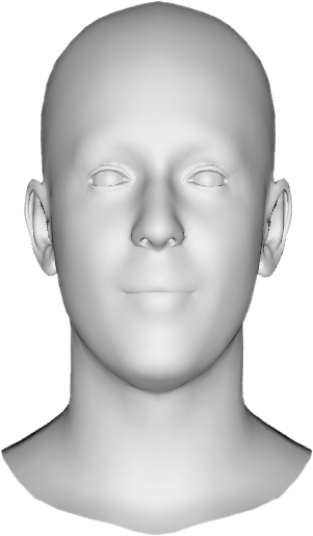} &
\includegraphics[scale=0.18]{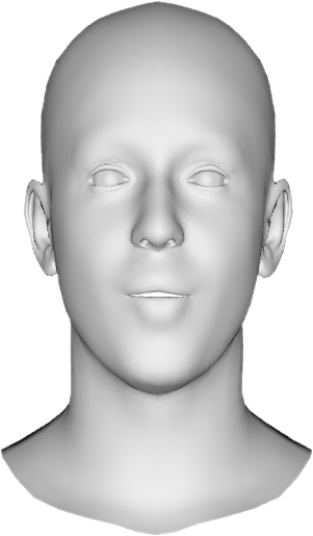} &
\includegraphics[scale=0.18]{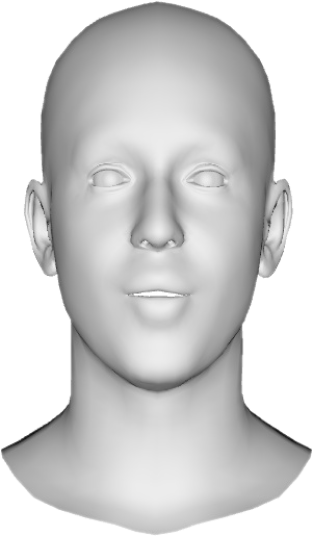} &
\includegraphics[scale=0.18]{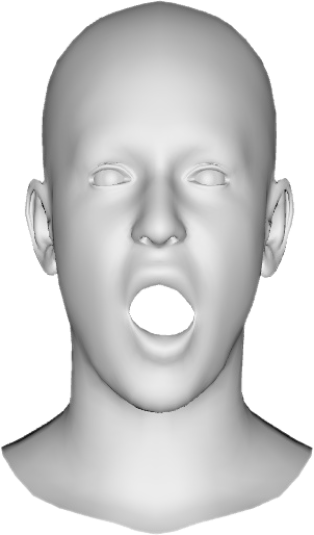} \\
\raisebox{5\height}{ ProbTalk3D} &
\includegraphics[scale=0.18]{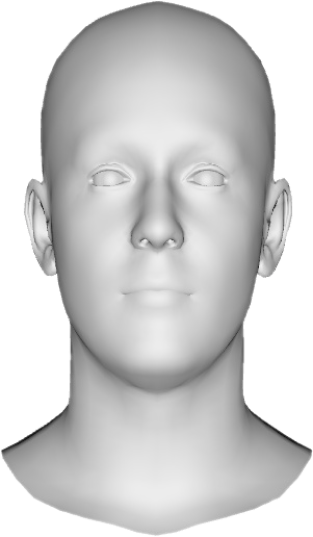} &
\includegraphics[scale=0.18]{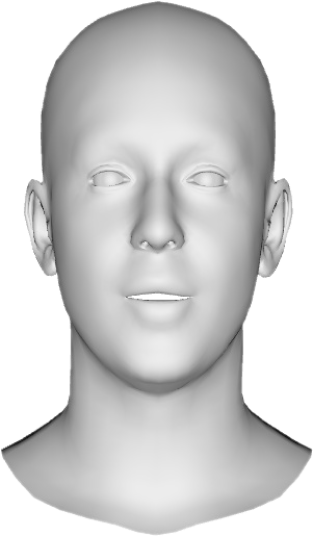} &
\includegraphics[scale=0.18]{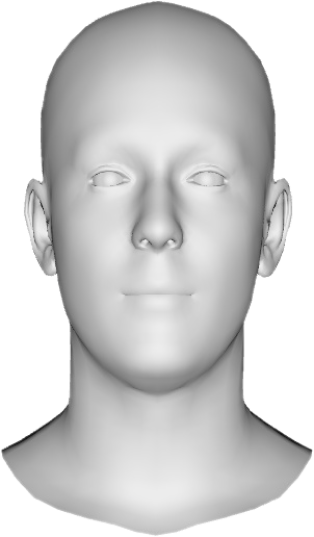} &
\includegraphics[scale=0.18]{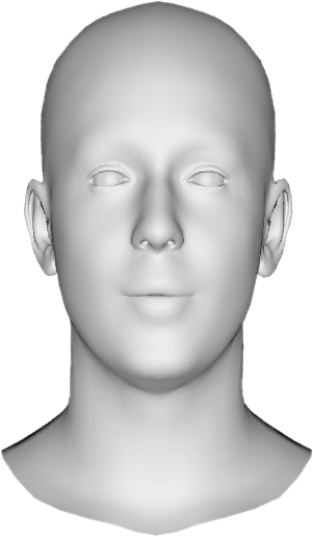} &
\includegraphics[scale=0.18]{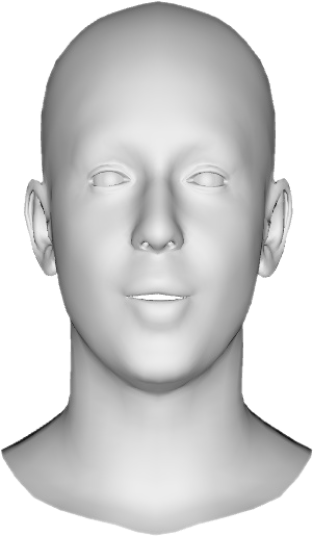} &
\includegraphics[scale=0.18]{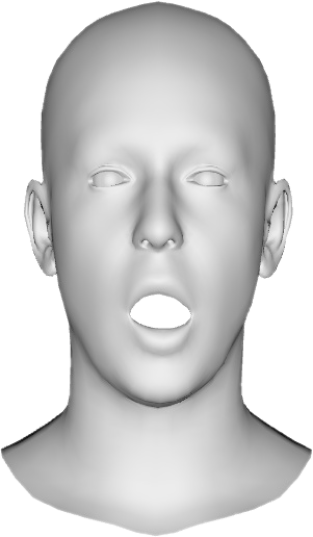} \\
\end{tabular}
}
\caption{Visual comparison of generated facial animations by different models together with ground truth (GT).}
\label{fig:quality_eva}
\Description{Visual comparisons of generated facial motions by different models}
\end{figure}

\paragraph{\textbf{Diversity}}
To assess diversity, we adopt the concept proposed in \cite{ren2023diffusion}. It was originally introduced for analyzing synthesized human motions and subsequently adapted for evaluating facial animations in DiffPoseTalk \cite{sun2023diffposetalk} and 3DiFACE \cite{thambiraja20233diface}. We aim to quantify diversity under an identical set of input conditions. Given $A$ audio inputs, we generate a set of facial animations comprising 10 samples for each audio, all guided by the same control signals. For the $i$-th audio input, we randomly sample two subsets, each containing $B$ samples from the generated animation set. In our analysis, $A=1909$ (i.e. test set audios) and $B=5$ (10 generated samples randomly split into two subsets $S_1$ and $S_2$ of 5 samples). Subsequently, we compute the average euclidean difference between the $j$-th sample within the two subsets. This process is repeated for all $A$ audio inputs, and the mean value is calculated as the final result. The diversity metric is formalized as:
\begin{equation} \label{equation_eva_div}
\text{Diversity} = \frac{1}{A \times B} \sum_{i=1}^{A} \sum_{j=1}^{B} \| (\hat{x}_{i,j} \in S_1) - (\hat{x}_{i,j} \in S_2)\|_2
\end{equation}

The comparative analysis among the models using the quantitative metrics is presented in Tab.\ref{tab:quant_result}. We can observe that ProbTalk3D scores significantly higher on the Diversity metric in comparison to VAE-based and Diffusion-based models. ProbTalk3D have similar values for MVE, LVE, MEE, and CE metrics in comparison to the VAE-based model while producing significantly better results than the original FaceDiffuser (with DDPM). The MVE, LVE, and FDD metrics which are calculated based on a single output sample demonstrate that the VAE-based method achieves slightly better results than ProbTalk3D, although pretty close. We argue that the MEE and CE metrics provide a better indication of lip movement accuracy for probabilistic models in comparison to LVE. VAE-based model and ProbTalk3D come closer to each other on these two metrics. FaceDiffuser-DDIM performs best on lip-sync related measures LVE, MEE and CE, while it performs worse than ProbTalk3D and VAE-based model for the whole face animation, upper face dynamics (MVE and FDD) and Diversity metrics. VAE-based model has the best FDD score followed by ProbTalk3D. The two models we present in this paper, ProbTalk3D and VAE-based model, produce the best results in terms of diversity and whole face animation related metrics while being comparable to FaceDiffuser on lip-related metrics. Non-deterministic methods produce better results than deterministic methods overall. 

The results points out to a trade-off between lip-sync accuracy, whole face animation, facial dynamics and Diversity metric. ProbTalk3D trades off some accuracy compared to the ground truth while achieving higher diversity. This accuracy-diversity trade-off for non-deterministic approaches was also observed and presented in \cite{yang2024probabilistic}. 

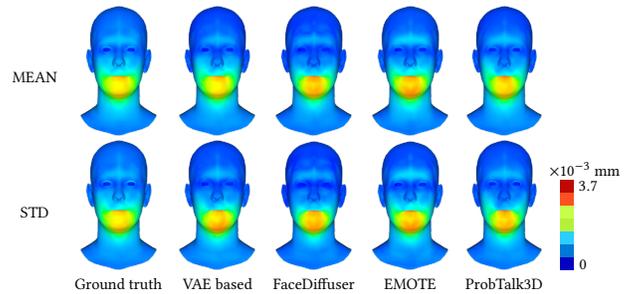
\begin{figure}[t]
\centering
\resizebox{0.9\linewidth}{!}{
\begin{tabular}{ccccccc}
\raisebox{5\height}{MEAN} &
\includegraphics[scale=0.18]{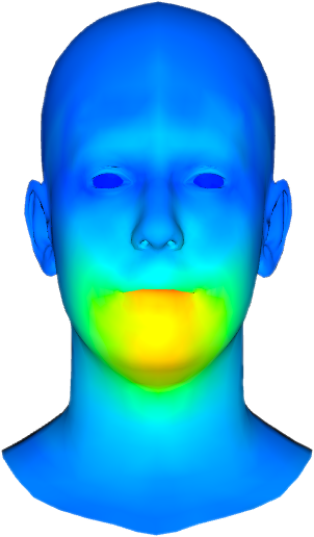} &
\includegraphics[scale=0.18]{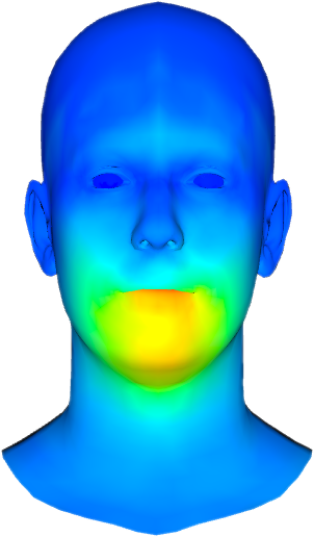} &
\includegraphics[scale=0.18]{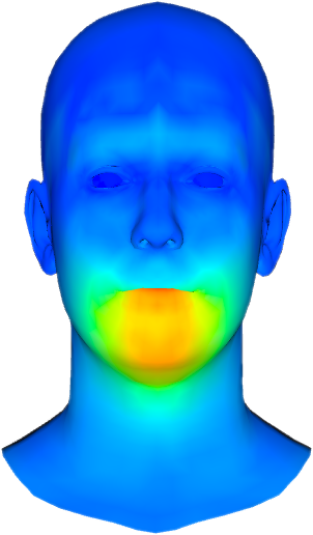} &
\includegraphics[scale=0.18]{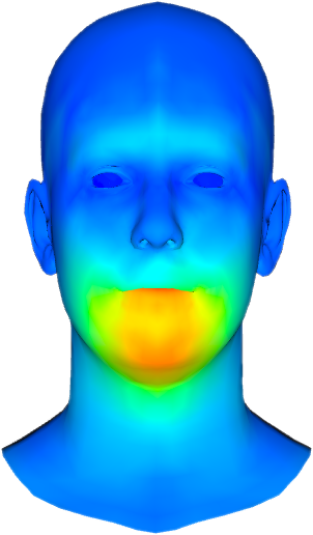} & 
\includegraphics[scale=0.18]{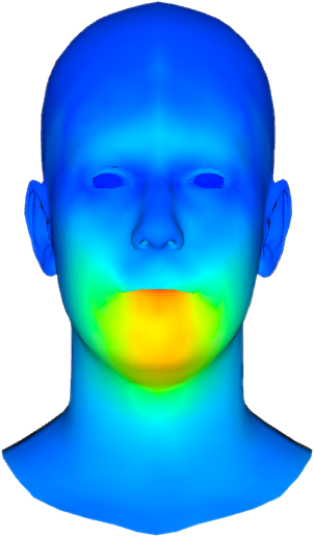} \\
\raisebox{5\height}{STD} &
\includegraphics[scale=0.18]{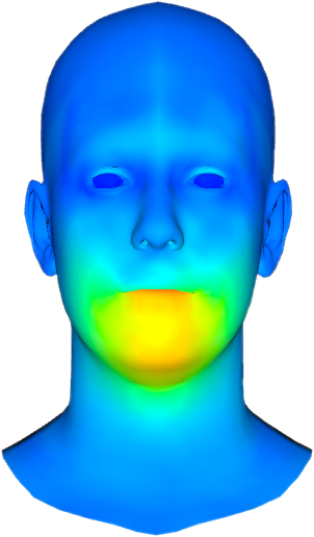} &
\includegraphics[scale=0.18]{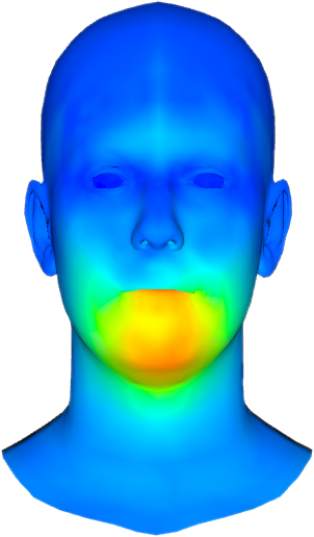} &
\includegraphics[scale=0.18]{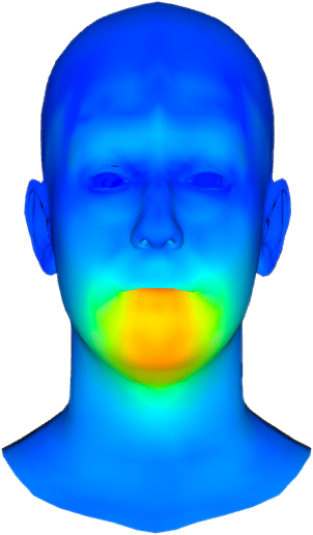} &
\includegraphics[scale=0.18]{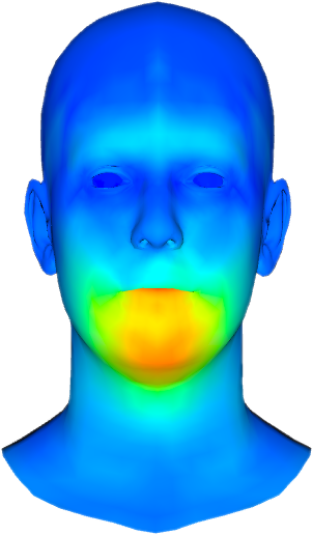} & 
\includegraphics[scale=0.18]{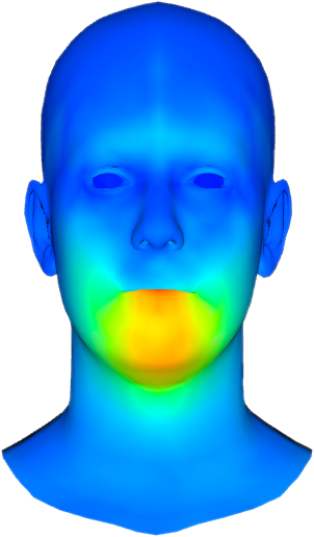} &
\begin{picture}(0,0)
    \put(0,0){\makebox(25,114){ $\times 10^{-3}$ mm}}
    \put(0,0){\makebox(29,95){ $3.7$}}
    \put(0,0){\includegraphics{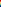}}
    \put(0,0){\makebox(23,6){ $0$}}
\end{picture} \\
& Ground truth & VAE based & FaceDiffuser  & EMOTE &  ProbTalk3D
\end{tabular}
}
\caption{Comparison using heatmap visualization of mean and standard deviation of generated animation by different models together with ground truth (GT) given audio sequence, uttering the sentence: ``He ate four extra eggs for breakfast". }
\label{fig:quality_heatmap}
\Description{Visual comparisons of generated facial motions using heatmaps.}
\end{figure}

\subsection{Qualitative Evaluation}
\label{sec:qualitative}
For qualitative assessment, we visually compare the generated animations of different models pronouncing specific syllables, focusing on assessing the quality of lip synchronization. The syllables are chosen to assess the model's capability in synthesizing diverse mouth shapes. This includes challenging sounds like the bilabial consonants /p/ and /m/, which require precise lip closure, as well as syllables that require a pout and sounds that demand an open-mouth posture. This quality judgement uses audio sequences that express a neutral emotion, selected from the test set of 3DMEAD. To compare our model against SOTAs, we include animations generated by the modified FaceDiffuser and the deterministic model, EMOTE. Note that we did not retrain EMOTE and utilized the publicly available trained model and inference script for motion generation. Fig.\ref{fig:quality_eva} showcases the results, with the ground truth provided as a reference. It can be seen from the figure that both ProbTalk3D and the VAE-based model closely resemble the ground truth, and achieve proper mouth closure. Inspecting the results of FaceDiffuser and EMOTE, we observe similar results indicating that our model performs comparable or better with respect to the recent non-deterministic and deterministic models.

\begin{figure}[t]
\centering
\resizebox{0.99\linewidth}{!}{
\begin{tabular}{ccccccc}
 & \textcolor{orange}{Wi}ll & \textcolor{orange}{yo}u & te\textcolor{orange}{ll} & \textcolor{orange}{me} & \textcolor{orange}{wh}- & \textcolor{orange}{y}? \\
\raisebox{6\height}{\footnotesize Neutral} &
\includegraphics[scale=0.18]{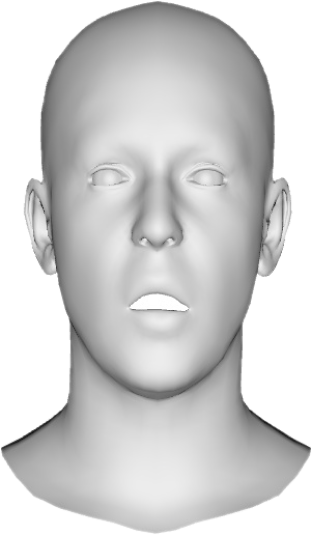} &
\includegraphics[scale=0.18]{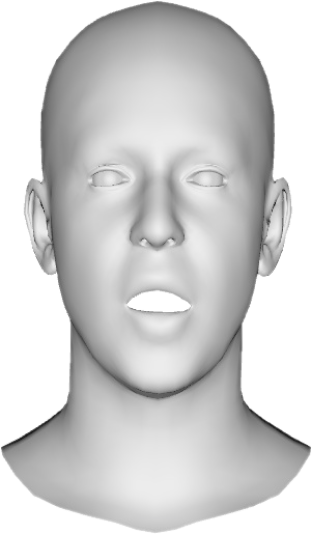} &
\includegraphics[scale=0.18]{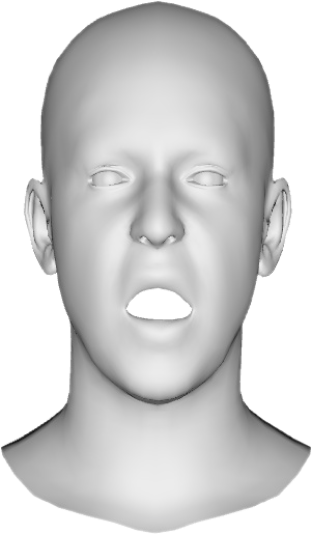} &
\includegraphics[scale=0.18]{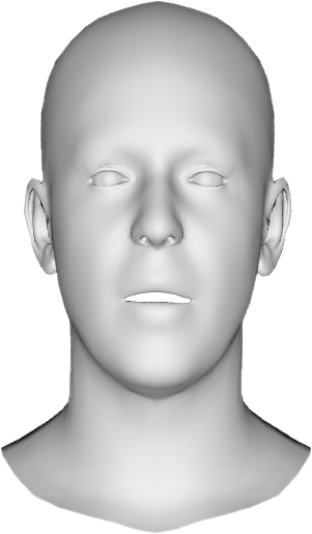} &
\includegraphics[scale=0.18]{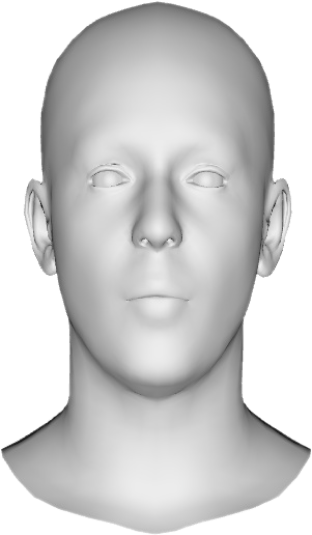} &
\includegraphics[scale=0.18]{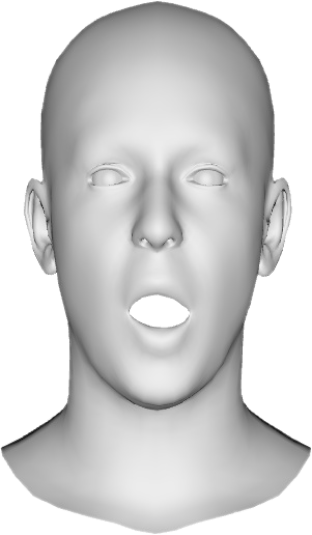} \\
\raisebox{6\height}{\footnotesize Happy} &
\includegraphics[scale=0.18]{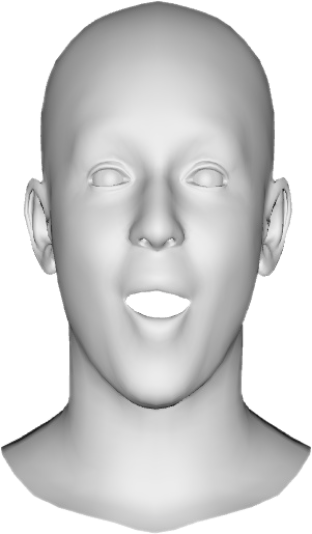} &
\includegraphics[scale=0.18]{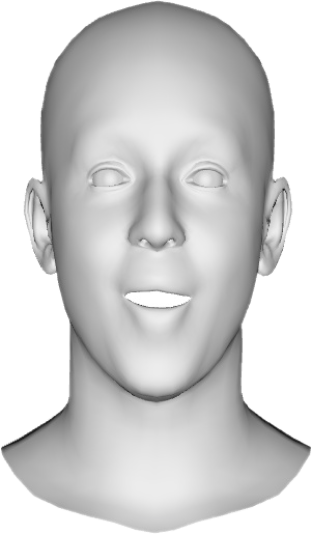} &
\includegraphics[scale=0.18]{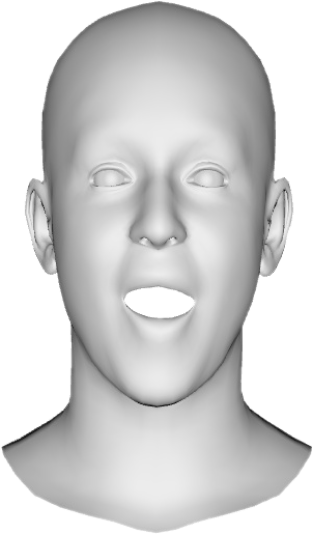} &
\includegraphics[scale=0.18]{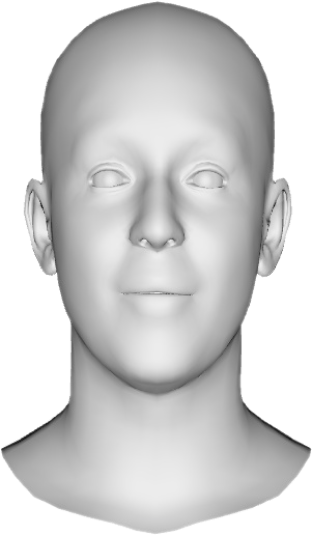} &
\includegraphics[scale=0.18]{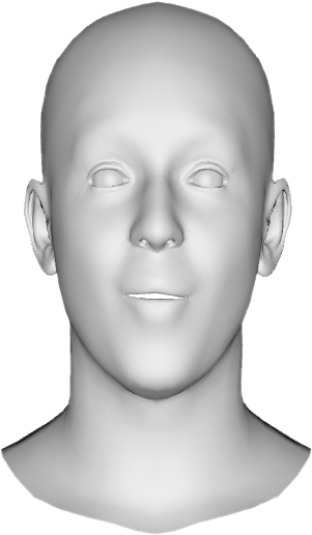} &
\includegraphics[scale=0.18]{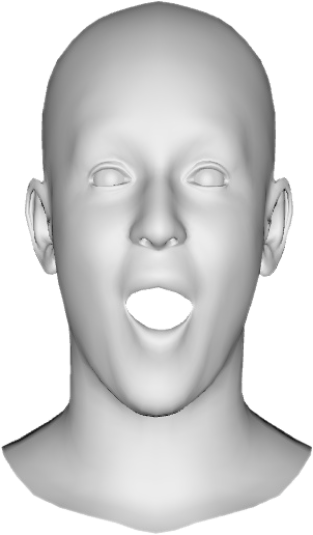} \\
\raisebox{6\height}{\footnotesize Sad} &
\includegraphics[scale=0.18]{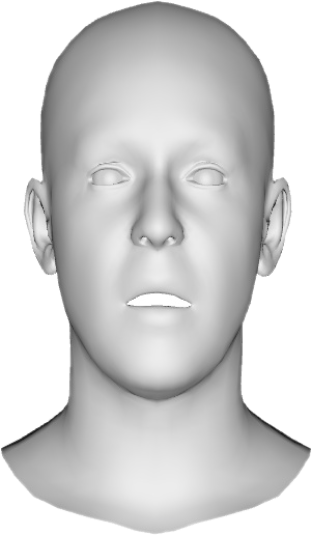} &
\includegraphics[scale=0.18]{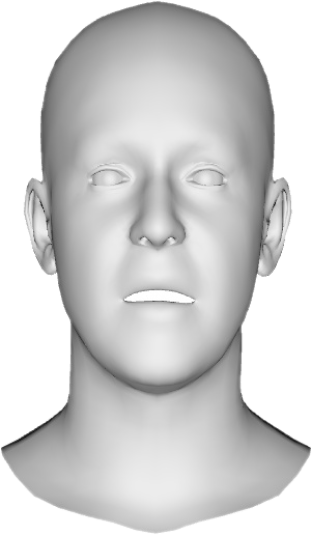} &
\includegraphics[scale=0.18]{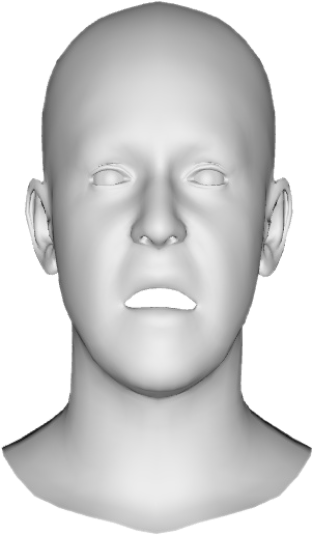} &
\includegraphics[scale=0.18]{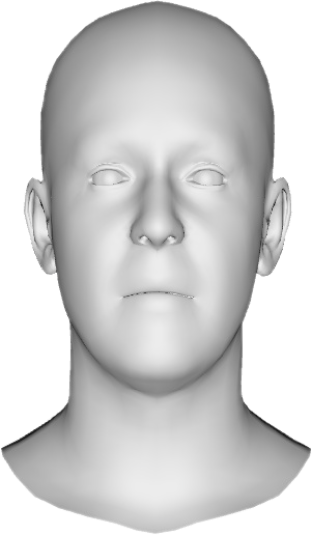} &
\includegraphics[scale=0.18]{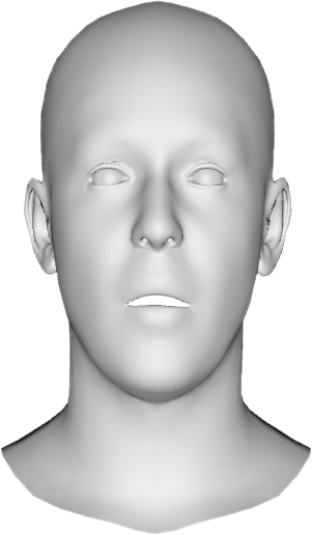} &
\includegraphics[scale=0.18]{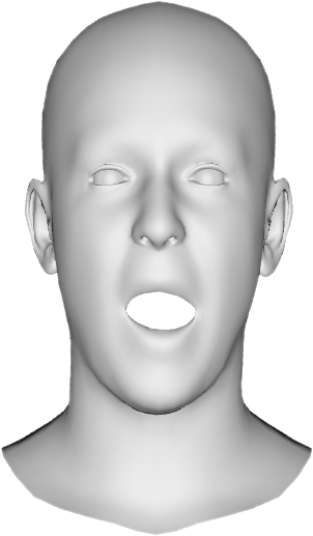} \\
\raisebox{6\height}{\footnotesize Surprised} &
\includegraphics[scale=0.18]{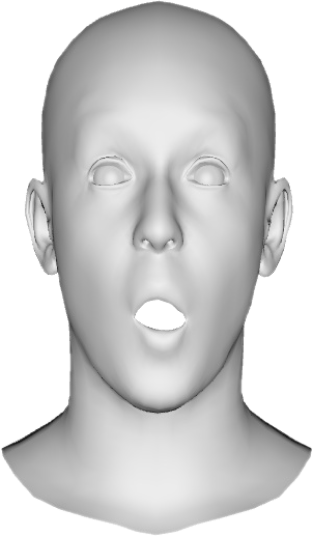} &
\includegraphics[scale=0.18]{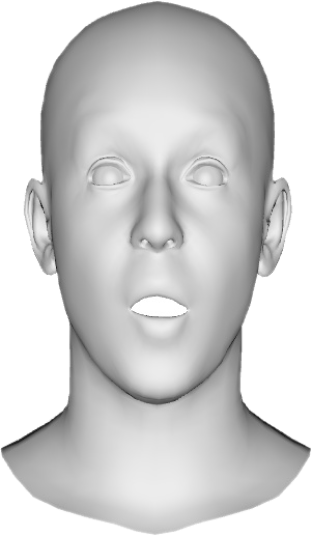} &
\includegraphics[scale=0.18]{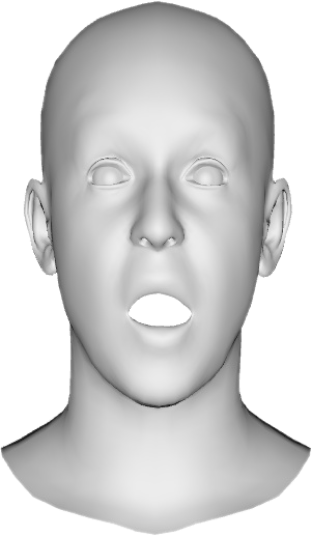} &
\includegraphics[scale=0.18]{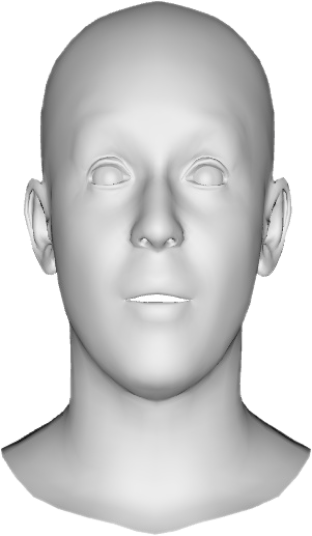} &
\includegraphics[scale=0.18]{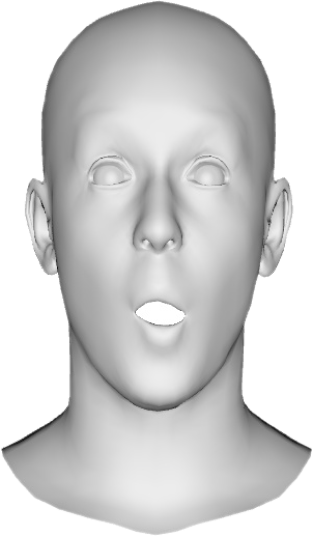} &
\includegraphics[scale=0.18]{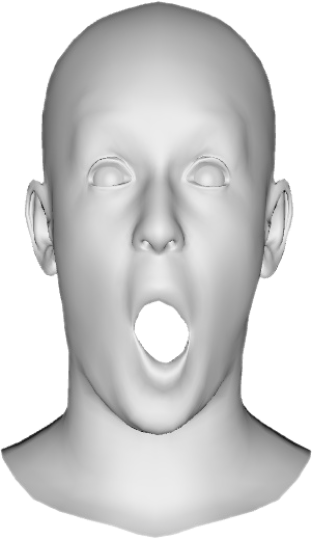} \\
\raisebox{6\height}{\footnotesize Fear} &
\includegraphics[scale=0.18]{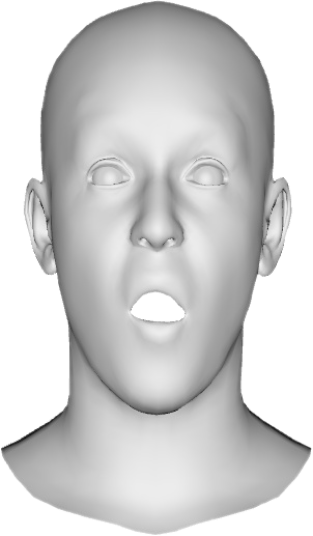} &
\includegraphics[scale=0.18]{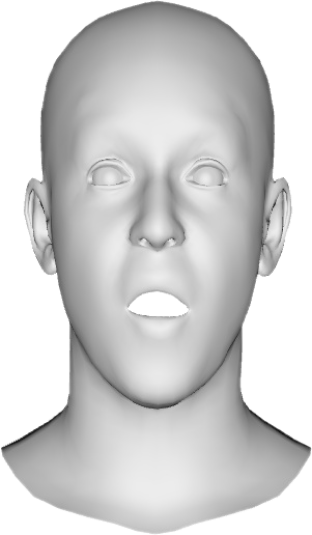} &
\includegraphics[scale=0.18]{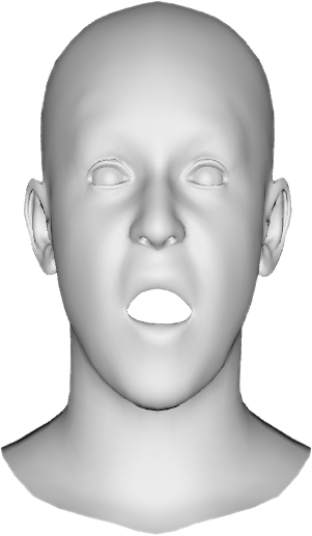} &
\includegraphics[scale=0.18]{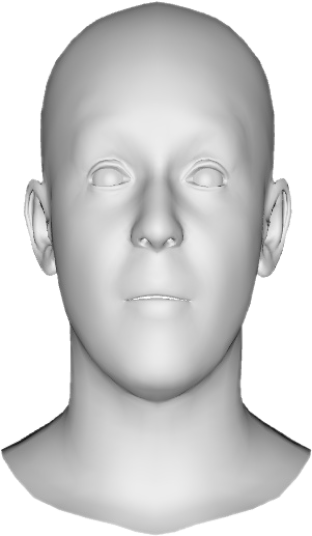} &
\includegraphics[scale=0.18]{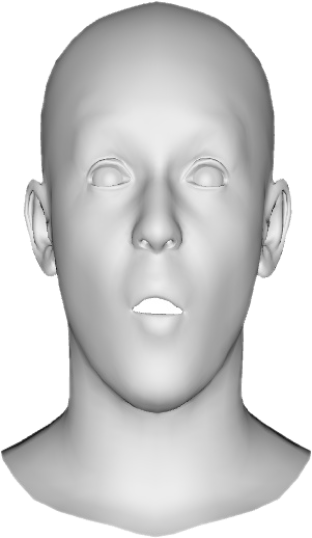} &
\includegraphics[scale=0.18]{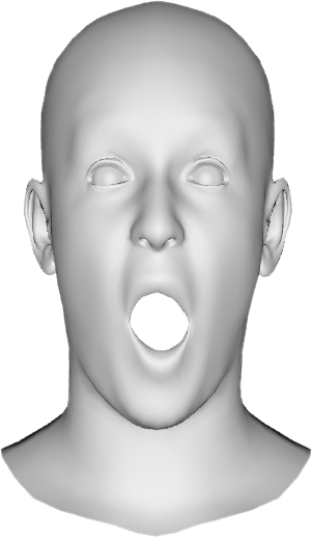} \\
\raisebox{6\height}{\footnotesize Disgusted} &
\includegraphics[scale=0.18]{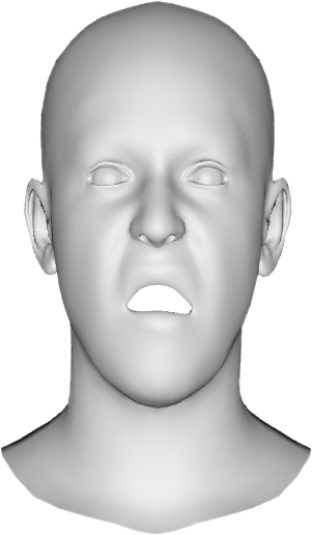} &
\includegraphics[scale=0.18]{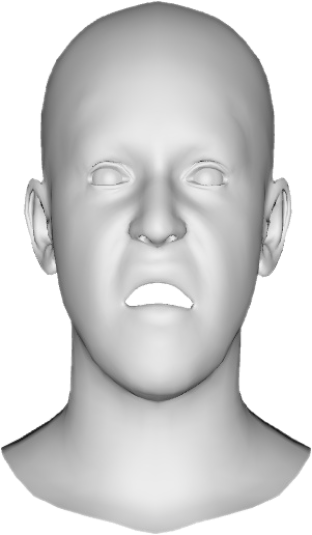} &
\includegraphics[scale=0.18]{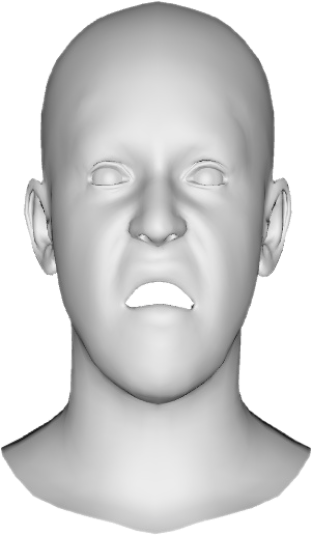} &
\includegraphics[scale=0.18]{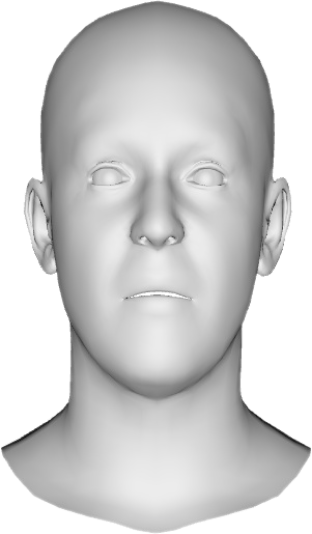} &
\includegraphics[scale=0.18]{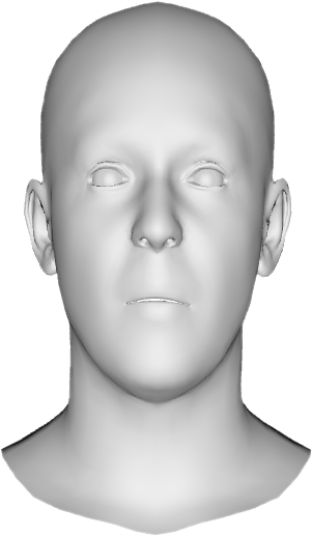} &
\includegraphics[scale=0.18]{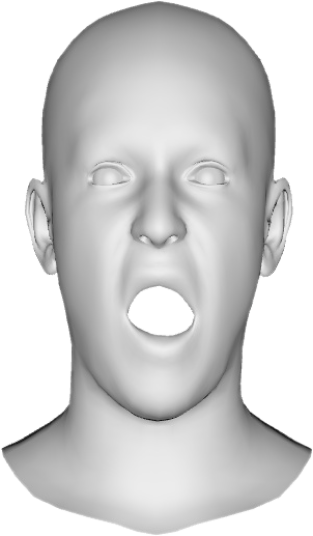} \\
\raisebox{6\height}{\footnotesize Angry} &
\includegraphics[scale=0.18]{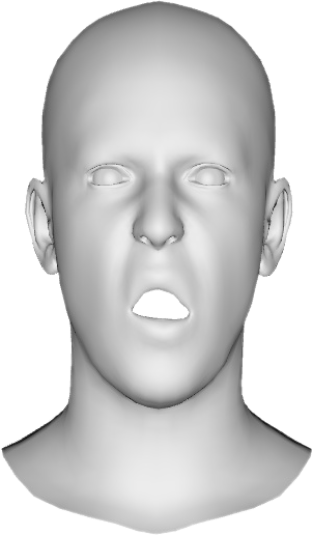} &
\includegraphics[scale=0.18]{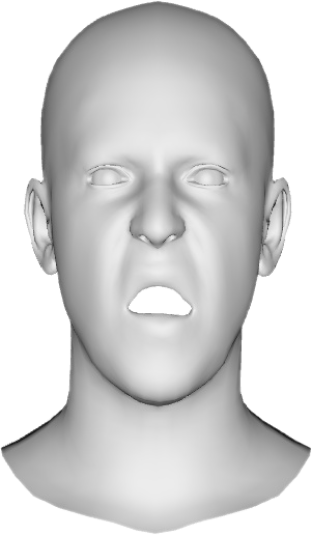} &
\includegraphics[scale=0.18]{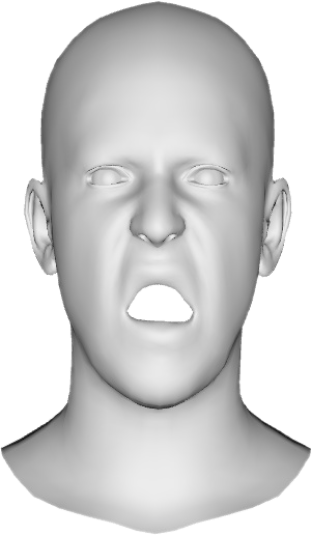} &
\includegraphics[scale=0.18]{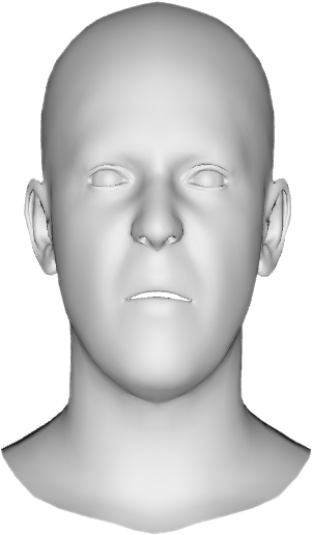} &
\includegraphics[scale=0.18]{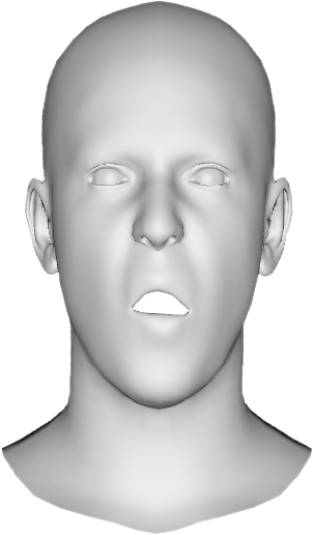} &
\includegraphics[scale=0.18]{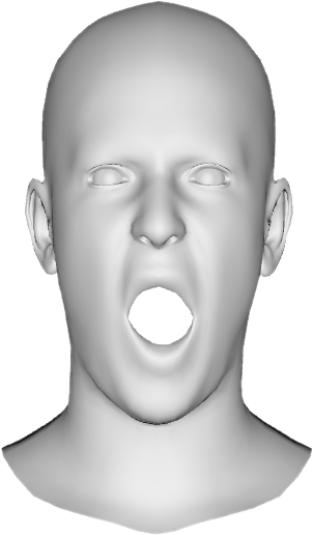} \\
\raisebox{6\height}{\footnotesize Contempt} &
\includegraphics[scale=0.18]{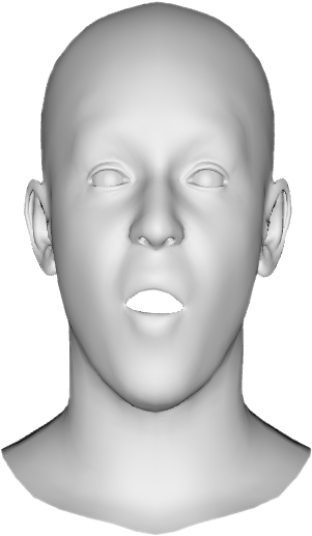} &
\includegraphics[scale=0.18]{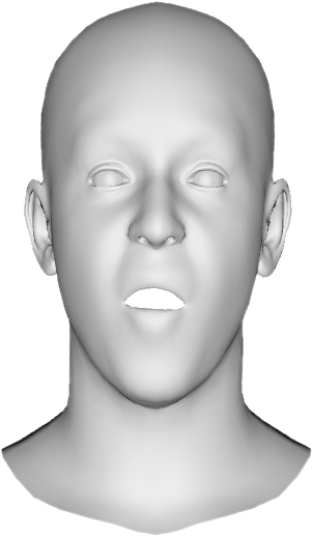} &
\includegraphics[scale=0.18]{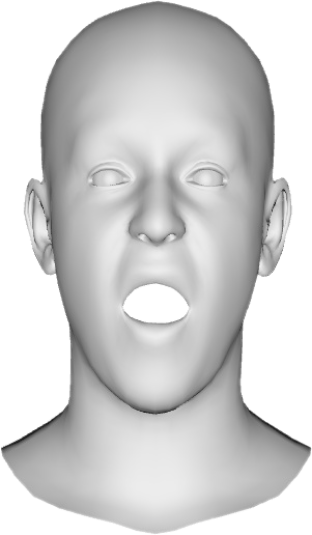} &
\includegraphics[scale=0.18]{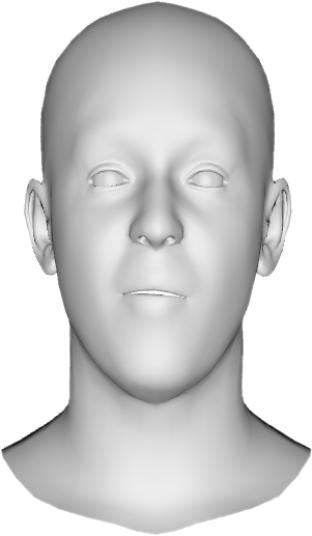} &
\includegraphics[scale=0.18]{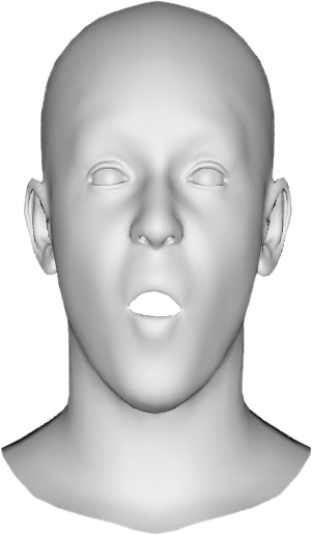} &
\includegraphics[scale=0.18]{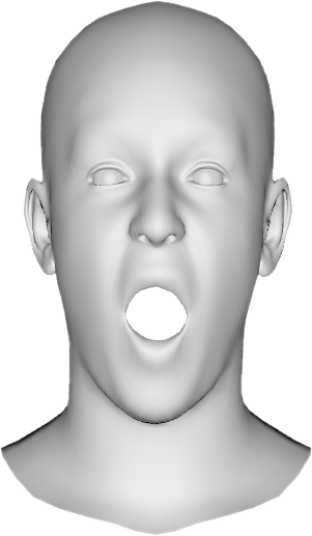} \\
\end{tabular}
}
\caption{Visual results of facial animations with different driving emotions generated by ProbTalk3D for an audio sequence uttering- \textit{``Will you tell me why?"}. Each row here showcases some key visual frames corresponding to the specific emotion controlled generated animation.}
\label{fig:quality_emo_VQVAE}
\Description{Emotion Generation}
\end{figure}

To illustrate the capability of our model in generating a wide range of facial movements, we adopt the approach used in \cite{xing2023codetalker} to present facial motion dynamics through heatmap visualization. We calculate the temporal statistics of adjacent-frame facial motions within a sequence, represented by the mean and standard deviation of the $\mathcal{L}_2$ distance between frames. A higher mean indicates more frequent facial movements, described by warmer colors in Figure \ref{fig:quality_heatmap}. Similarly, a higher standard deviation represents richer variations in facial dynamics. Fig.\ref{fig:quality_heatmap} demonstrates that both ProbTalk3D and the VAE-based mode can produce animations with diverse facial movements in the mouth region and the forehead. The predicted motions closely resemble the dynamics of the ground truth, showcasing the accuracy of our approaches. Furthermore, we present Fig.\ref{fig:quality_emo_VQVAE} to demonstrate our model's capability to generate emotion-controlled facial animation. Evidently, the distinction between emotions is clear and perceptually aligns with the semantic labels.

\subsection{Perceptual User Study}
We conduct A/B testing to compare our proposed model's results against ground truth and results from SOTA models- FaceDiffuser-DDIM (non-deterministic - trained on 3DMEAD with emotion control) and EMOTE (i.e. deterministic - trained on 3DMEAD with emotion control using the publicly available model). These SOTAs have previously demonstrated superior performance compared to several earlier works. For the user study, we generate 32 videos using each model. These videos cover 8 emotions and different speaking styles, with 4 video sequences generated for each emotion. For the 7 emotions other than neutral, we create 2 videos with medium intensity and 2 with high intensity. Videos under a specific emotion are synthesized using 2 audio sequences from 3DMEAD (with identities unseen during the second stage of training) and 2 in-the-wild audio samples from the VoxMovies \cite{brown2021playing} dataset for generalizability. Additionally, we generate another set of 32 videos using our model to compare with the ground truth belonging to the unseen test-set of 3DMEAD. The survey is set to randomly choose 1 video pair within 4 pairs for each emotion. This results in a total of 24 video pairs: 8 pairs each comparing our model with- (i) the ground truth, (ii) FaceDiffuser and (iii) EMOTE. 

We use Qualtrics \cite{qualtrics} as the survey tool and recruit participants through Prolific \cite{Prolific}, ensuring proper remuneration. Participants are queried about lip synchronization, realism, and emotional expressivity after viewing each video pair, requiring them to select the one they perceive as better. In total, 73 responses are collected, with 4 responses discarded due to failing the attention test, resulting in 69 valid responses. The result of the perceptual study is reported in Tab.\ref{tab:user_study}, demonstrating that our model's output is generally less preferred than the ground truth across lip synchronization, realism, and emotional expression. This is expected and in line with other recent works. However, our model outperforms the competitors in all three aspects. The user study demonstrates the superiority of our model over the competitors, showcasing its capability to generate animations with good lip synchronization, high overall face realism, and superior emotional expressivity. More details about the user study can be found in the supplementary material. 

\begin{table*}[]
\centering
\begin{tabular}{cccc}
& \footnotesize \textbf{Lip Sync} & \footnotesize \textbf{Realism} & \footnotesize \textbf {Emotional Expression} \\
\raisebox{1.2ex}{
\centering\arraybackslash\begin{tabular}{@{}c@{}} 
{ \textcolor{darkcolumbiablue}{Ours} vs.} { \textcolor{darkgreen}{GT}} \end{tabular}} &
\begin{tikzpicture}
    \fill[columbiablue] (0,0) rectangle (1.572,0.5); 
    \fill[mossgreen] (1.572,0) rectangle (4,0.5); 
    \node at (0.79,0.25) {\footnotesize{39.31\%}};
    \node at (2.79,0.25) {\footnotesize{60.69\%}};
\end{tikzpicture} & 
\begin{tikzpicture}
    \fill[columbiablue] (0,0) rectangle (1.703,0.5); 
    \fill[mossgreen] (1.703,0) rectangle (4,0.5); 
    \node at (0.85,0.25) {\footnotesize{42.57\%}};
    \node at (2.85,0.25) {\footnotesize{57.43\%}};
\end{tikzpicture} &
\begin{tikzpicture}
    \fill[columbiablue] (0,0) rectangle (1.536,0.5); 
    \fill[mossgreen] (1.536,0) rectangle (4,0.5); 
    \node at (0.77,0.25) {\footnotesize{38.41\%}};
    \node at (2.77,0.25) {\footnotesize{61.59\%}};
\end{tikzpicture} \\ 
\raisebox{1.2ex}{
\centering\arraybackslash\begin{tabular}{@{}c@{}} 
{ \textcolor{darkcolumbiablue}{Ours} vs.} { \textcolor{lightsalmonpink}{FaceDiffuser}} \end{tabular}} &
\begin{tikzpicture}
    \fill[columbiablue] (0,0) rectangle (2.370,0.5); 
    \fill[lightsalmonpink] (2.370,0) rectangle (4,0.5); 
    \node at (1.18,0.25) {\footnotesize{59.24\%}};
    \node at (3.18,0.25) {\footnotesize{40.76\%}};
\end{tikzpicture} & 
\begin{tikzpicture}
    \fill[columbiablue] (0,0) rectangle (2.377,0.5); 
    \fill[lightsalmonpink] (2.377,0) rectangle (4,0.5); 
    \node at (1.19,0.25) {\footnotesize{59.42\%}};
    \node at (3.19,0.25) {\footnotesize{40.58\%}};
\end{tikzpicture} &
\begin{tikzpicture}
    \fill[columbiablue] (0,0) rectangle (2.406,0.5); 
    \fill[lightsalmonpink] (2.406,0) rectangle (4,0.5); 
    \node at (1.20,0.25) {\footnotesize{60.14\%}};
    \node at (3.20,0.25) {\footnotesize{39.86\%}};
\end{tikzpicture} \\
\raisebox{1.2ex}{
\centering\arraybackslash\begin{tabular}{@{}c@{}} 
{ \textcolor{darkcolumbiablue}{Ours} vs.} { \textcolor{darkgold}{EMOTE}} \end{tabular}} &
\begin{tikzpicture}
    \fill[columbiablue] (0,0) rectangle (2.239,0.5); 
    \fill[gold] (2.239,0) rectangle (4,0.5); 
    \node at (1.12,0.25) {\footnotesize{55.98\%}};
    \node at (3.12,0.25) {\footnotesize{44.02\%}};
\end{tikzpicture} &
\begin{tikzpicture}
    \fill[columbiablue] (0,0) rectangle (2.152,0.5); 
    \fill[gold] (2.152,0) rectangle (4,0.5); 
    \node at (1.08,0.25) {\footnotesize{53.80\%}};
    \node at (3.08,0.25) {\footnotesize{46.20\%}};
\end{tikzpicture} &
\begin{tikzpicture}
    \fill[columbiablue] (0,0) rectangle (2.275,0.5); 
    \fill[gold] (2.275,0) rectangle (4,0.5); 
    \node at (1.14,0.25) {\footnotesize{56.88\%}};
    \node at (3.14,0.25) {\footnotesize{43.12\%}};
\end{tikzpicture} \\
\vspace{1pt}
\end{tabular}
\caption{We conduct A/B testing to assess our model's perceived quality in terms of lip synchronization, realism, and emotional expressivity. We compare against the ground truth, FaceDiffuser, and EMOTE where the result reports the percentage of times each model's output (or the ground truth animation, in case of Ours vs. GT) is perceived better.}
\label{tab:user_study}
\end{table*}

\subsection{Ablation Study}
\label{sec:ablation_study}
\begin{figure}[t]
\centering
\resizebox{0.8\linewidth}{!}{
\begin{tabular}{cccc}
\raisebox{5\height}{\footnotesize w/ emotion control} &
\includegraphics[scale=0.18]{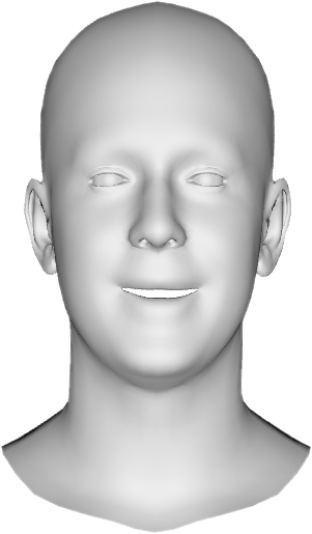} &
\includegraphics[scale=0.18]{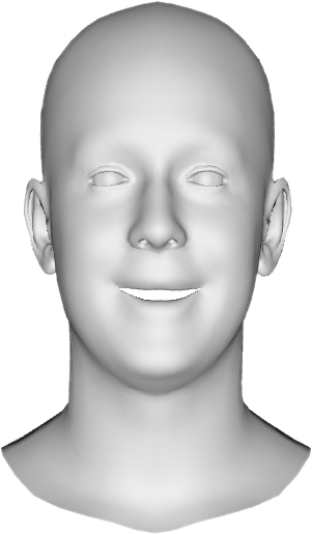} &
\multirow{2}{*}[12pt]{\includegraphics[scale=0.18]{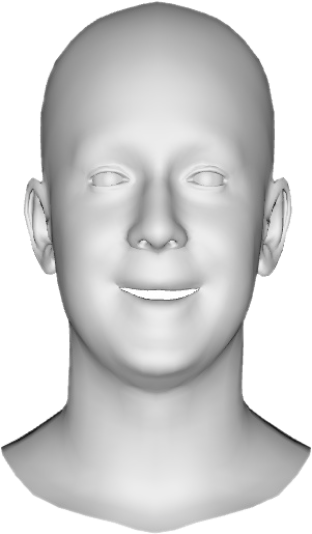}}\\
\raisebox{5\height}{\footnotesize w/o emotion control} & \includegraphics[scale=0.18]{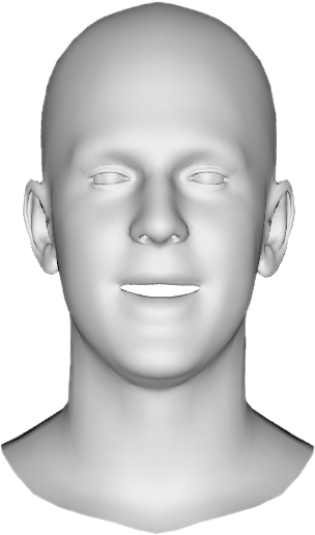} &
\includegraphics[scale=0.18]{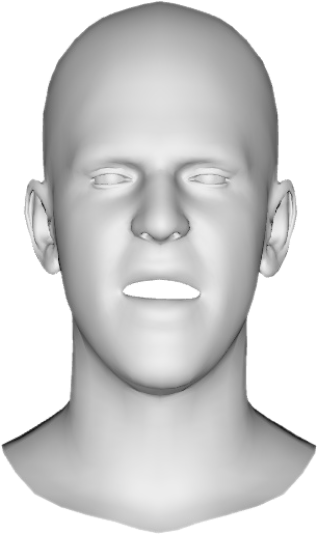} \\
 & \footnotesize VAE based & \footnotesize Proposed & \footnotesize GT
\end{tabular}}
\caption{Visual comparison against ground truth (GT) frame using frames generated by proposed and VAE-based variant models trained with and without emotion control.}
\label{fig:quality_noemo}
\Description{Ablation figure}
\end{figure}

We conduct an ablation study to understand whether emotion control enhances synthesized animation quality. We remove the Style Vector input, $\mathcal{C}$, and instead, learn a general representation that is not specific to any of the style vector attributes. The remaining training settings stay identical to the proposed approach. The objective metrics related to the ablation study are provided in Tab.\ref{tab:quant_result}. When emotion control is excluded, ProbTalk3D outperforms the VAE variant across all evaluation metrics. The VAE variant without emotion control shows worse reconstruction quality but a slightly better Diversity value than the emotion control-enabled VAE model. ProbTalk3D without emotion control achieves slightly better results in LVE, MEE, and CE metrics in comparison to the version with emotion control. Fig.\ref{fig:quality_noemo} displays a frame of the synthesized motion generated by models with and without emotion control, given the same audio input. The ground truth frame is included for reference. We notice that without emotion control, the models produce less expressive outputs or fail to accurately convey the intended emotion.

\section{Discussion and Future Work}
Our model is limited to generate facial animations conditioned on 8 basic emotions and 3 discrete emotion intensities as per ground truth annotations of the dataset. However, human emotion is much more detailed and richer to be controlled by pre-defined categories. We believe that by combining our model with textual descriptions would enable us to learn and control the generation of richer emotions rather than relying solely on one-hot vectors for style embedding, similar to \cite{Ao2023GestureDiffuCLIP} which was applied in the body motion generation domain. Furthermore, due to the 3DMEAD dataset, our model shares similar limitations as EMOTE \cite{emote} as presented in their work that include- (i) not being able to generate eye blinks, (ii) absence of mouth cavity, teeth and tongue animation that effect perception of speech animation (iii) visual artefacts for high-frequency speech as the reconstructed visual data is of low frequency.  

Similar to our model ProbTalk3D and the VAE variant, we believe a 2-stage diffusion-based model, utilizing the more recent conditional diffusion \cite{park2023said} or latent diffusion \cite{chen2023executing} approaches might prove to be fruitful. Latent diffusion methods have demonstrated success in image generation and are applied in recent models for human motion synthesis \cite{chen2023executing} for enhanced diversity. More experiments are needed to effectively leverage diffusion based approaches for 3D facial animation synthesis. Moreover, additional datasets can be used for training and analysis to validate our model's generalizability and 4D datasets can be employed for enhanced realism.

Intuitively, the animation produced by a generative model should resemble human-like facial motion, meaning the diversity between generations should be close to that of the ground truth. If a model can generate a wide range of diverse samples, but they lack meaning, it does not necessarily indicate superior performance. However, to the best of our knowledge, in contrast with body animation datasets \cite{surveybodymotion}, no existing facial animation datasets exhibit this kind of natural diversity. Current datasets that include different ways of expressing the same sentence often involve performers being directed to convey specific emotions. To obtain ground truth diversity, we need multiple performances under the same style condition. With this information in the dataset, we can evaluate our models by assessing how closely their diversity matches as observed in ground truth. Future work can focus on constructing datasets that enable this diversity analysis. 

\section{Conclusion}
In this work, we propose ProbTalk3D, a novel non-deterministic approach for emotion controllable speech-driven 3D facial animation synthesis that outperforms most recent deterministic and non-deterministic methods. Our approach is based on VQ-VAE and trained in 2 stages. In the first stage, we learn a motion prior leveraging a VQ-VAE based motion autoencoder. In the second stage, we train our audio and emotion conditioned 3D facial animation synthesis network that takes advantage of the learned motion prior for generation. Extensive evaluations have been conducted to validate our model performance. Quantitative evaluation results demonstrate that our model achieves results that are comparable to state-of-the-arts while ensuring a wider range of diverse yet high-quality acceptable animation outputs. Furthermore, qualitative comparisons with the non-deterministic model- FaceDiffuser\cite{FaceDiffuser_Stan_MIG2023} and deterministic model- EMOTE\cite{emote} show that our approach perform better against state-of-the-art models while being less computationally complex and more efficient. The perceptual user study provides further evidence that our model is preferred to both FaceDiffuser and EMOTE on lip synchronization, realism, and emotional expressivity ratings. Our work highlights the necessity of non-deterministic methodologies for generative 3D facial animation by introducing a novel probabilistic model that is capable of generating high-quality yet diverse animations. We hope our findings will inspire new discussions and research directions in generative 3D facial animation landscape.

\begin{acks}
We thank the authors of FaceFormer, CodeTalker, EMOTE and MEAD dataset for making their codebases and datasets available. 
\end{acks}

\bibliographystyle{ACM-Reference-Format}
\bibliography{main}


\begin{thebibliography}{69}


\ifx \showCODEN    \undefined \def \showCODEN     #1{\unskip}     \fi
\ifx \showDOI      \undefined \def \showDOI       #1{#1}\fi
\ifx \showISBNx    \undefined \def \showISBNx     #1{\unskip}     \fi
\ifx \showISBNxiii \undefined \def \showISBNxiii  #1{\unskip}     \fi
\ifx \showISSN     \undefined \def \showISSN      #1{\unskip}     \fi
\ifx \showLCCN     \undefined \def \showLCCN      #1{\unskip}     \fi
\ifx \shownote     \undefined \def \shownote      #1{#1}          \fi
\ifx \showarticletitle \undefined \def \showarticletitle #1{#1}   \fi
\ifx \showURL      \undefined \def \showURL       {\relax}        \fi
\providecommand\bibfield[2]{#2}
\providecommand\bibinfo[2]{#2}
\providecommand\natexlab[1]{#1}
\providecommand\showeprint[2][]{arXiv:#2}

\bibitem[Alexanderson et~al\mbox{.}(2023)]%
        {alexanderson2023listen}
\bibfield{author}{\bibinfo{person}{Simon Alexanderson}, \bibinfo{person}{Rajmund Nagy}, \bibinfo{person}{Jonas Beskow}, {and} \bibinfo{person}{Gustav~Eje Henter}.} \bibinfo{year}{2023}\natexlab{}.
\newblock \showarticletitle{Listen, Denoise, Action! Audio-Driven Motion Synthesis with Diffusion Models}.
\newblock \bibinfo{journal}{\emph{ACM Trans. Graph.}} \bibinfo{volume}{42}, \bibinfo{number}{4} (\bibinfo{year}{2023}), \bibinfo{pages}{1--20}.
\newblock
\urldef\tempurl%
\url{https://doi.org/10.1145/3592458}
\showDOI{\tempurl}


\bibitem[Aneja et~al\mbox{.}(2024a)]%
        {FaceTalk_Aneja_2024_CVPR}
\bibfield{author}{\bibinfo{person}{Shivangi Aneja}, \bibinfo{person}{Justus Thies}, \bibinfo{person}{Angela Dai}, {and} \bibinfo{person}{Matthias Nie{\ss}ner}.} \bibinfo{year}{2024}\natexlab{a}.
\newblock \showarticletitle{FaceTalk: Audio-Driven Motion Diffusion for Neural Parametric Head Models}. In \bibinfo{booktitle}{\emph{Proceedings of the IEEE/CVF Conference on Computer Vision and Pattern Recognition (CVPR)}}. \bibinfo{pages}{21263--21273}.
\newblock


\bibitem[Aneja et~al\mbox{.}(2024b)]%
        {aneja2023facetalk}
\bibfield{author}{\bibinfo{person}{Shivangi Aneja}, \bibinfo{person}{Justus Thies}, \bibinfo{person}{Angela Dai}, {and} \bibinfo{person}{Matthias Nießner}.} \bibinfo{year}{2024}\natexlab{b}.
\newblock \showarticletitle{FaceTalk: Audio-Driven Motion Diffusion for Neural Parametric Head Models}. In \bibinfo{booktitle}{\emph{Proc. IEEE Conf. on Computer Vision and Pattern Recognition (CVPR)}}.
\newblock


\bibitem[Ao et~al\mbox{.}(2023)]%
        {Ao2023GestureDiffuCLIP}
\bibfield{author}{\bibinfo{person}{Tenglong Ao}, \bibinfo{person}{Zeyi Zhang}, {and} \bibinfo{person}{Libin Liu}.} \bibinfo{year}{2023}\natexlab{}.
\newblock \showarticletitle{GestureDiffuCLIP: Gesture Diffusion Model with CLIP Latents}.
\newblock \bibinfo{journal}{\emph{ACM Trans. Graph.}} (\bibinfo{year}{2023}), \bibinfo{numpages}{18}~pages.
\newblock
\urldef\tempurl%
\url{https://doi.org/10.1145/3592097}
\showDOI{\tempurl}


\bibitem[Aylagas et~al\mbox{.}(2022)]%
        {Voice2FaceEA}
\bibfield{author}{\bibinfo{person}{Mónica~Villanueva Aylagas}, \bibinfo{person}{Héctor~Anadon Leon}, \bibinfo{person}{Mattias Teye}, {and} \bibinfo{person}{Konrad Tollmar}.} \bibinfo{year}{2022}\natexlab{}.
\newblock \showarticletitle{Voice2Face: Audio-driven Facial and Tongue Rig Animations with cVAEs}. In \bibinfo{booktitle}{\emph{EUROGRAPHICS SYMPOSIUM ON COMPUTER ANIMATION (SCA 2022}}.
\newblock


\bibitem[Brown et~al\mbox{.}(2021)]%
        {brown2021playing}
\bibfield{author}{\bibinfo{person}{Andrew Brown}, \bibinfo{person}{Jaesung Huh}, \bibinfo{person}{Arsha Nagrani}, \bibinfo{person}{Joon~Son Chung}, {and} \bibinfo{person}{Andrew Zisserman}.} \bibinfo{year}{2021}\natexlab{}.
\newblock \showarticletitle{Playing a part: Speaker verification at the movies}. In \bibinfo{booktitle}{\emph{ICASSP 2021-2021 IEEE International Conference on Acoustics, Speech and Signal Processing (ICASSP)}}. IEEE, \bibinfo{pages}{6174--6178}.
\newblock


\bibitem[Charalambous et~al\mbox{.}(2019)]%
        {Charalambous2019}
\bibfield{author}{\bibinfo{person}{Constantinos Charalambous}, \bibinfo{person}{Zerrin Yumak}, {and} \bibinfo{person}{A.F. van~der Stappen}.} \bibinfo{year}{2019}\natexlab{}.
\newblock \showarticletitle{Audio‐driven emotional speech animation for interactive virtual characters}.
\newblock \bibinfo{journal}{\emph{Computer Animation and Virtual Worlds}}  \bibinfo{volume}{30} (\bibinfo{year}{2019}).
\newblock


\bibitem[Chen et~al\mbox{.}(2024)]%
        {Diffsheg}
\bibfield{author}{\bibinfo{person}{Junming Chen}, \bibinfo{person}{Yunfei Liu}, \bibinfo{person}{Jianan Wang}, \bibinfo{person}{Ailing Zeng}, \bibinfo{person}{Yu Li}, {and} \bibinfo{person}{Qifeng Chen}.} \bibinfo{year}{2024}\natexlab{}.
\newblock \showarticletitle{DiffSHEG: A Diffusion-Based Approach for Real-Time Speech-driven Holistic 3D Expression and Gesture Generation}. In \bibinfo{booktitle}{\emph{Proceedings of the IEEE/CVF Conference on Computer Vision and Pattern Recognition (CVPR)}}. \bibinfo{pages}{7352--7361}.
\newblock


\bibitem[Chen et~al\mbox{.}(2023b)]%
        {chen2023diffusiontalker}
\bibfield{author}{\bibinfo{person}{Peng Chen}, \bibinfo{person}{Xiaobao Wei}, \bibinfo{person}{Ming Lu}, \bibinfo{person}{Yitong Zhu}, \bibinfo{person}{Naiming Yao}, \bibinfo{person}{Xingyu Xiao}, {and} \bibinfo{person}{Hui Chen}.} \bibinfo{year}{2023}\natexlab{b}.
\newblock \showarticletitle{DiffusionTalker: Personalization and Acceleration for Speech-Driven 3D Face Diffuser}.
\newblock \bibinfo{journal}{\emph{arXiv preprint arXiv:2311.16565}} (\bibinfo{year}{2023}).
\newblock


\bibitem[Chen et~al\mbox{.}(2023a)]%
        {chen2023executing}
\bibfield{author}{\bibinfo{person}{Xin Chen}, \bibinfo{person}{Biao Jiang}, \bibinfo{person}{Wen Liu}, \bibinfo{person}{Zilong Huang}, \bibinfo{person}{Bin Fu}, \bibinfo{person}{Tao Chen}, {and} \bibinfo{person}{Gang Yu}.} \bibinfo{year}{2023}\natexlab{a}.
\newblock \showarticletitle{Executing your commands via motion diffusion in latent space}. In \bibinfo{booktitle}{\emph{Proceedings of the IEEE/CVF Conference on Computer Vision and Pattern Recognition}}. \bibinfo{pages}{18000--18010}.
\newblock


\bibitem[Chhatre et~al\mbox{.}(2024)]%
        {Chhatre_2024_CVPR}
\bibfield{author}{\bibinfo{person}{Kiran Chhatre}, \bibinfo{person}{Radek Daněček}, \bibinfo{person}{Nikos Athanasiou}, \bibinfo{person}{Giorgio Becherini}, \bibinfo{person}{Christopher Peters}, \bibinfo{person}{Michael~J. Black}, {and} \bibinfo{person}{Timo Bolkart}.} \bibinfo{year}{2024}\natexlab{}.
\newblock \showarticletitle{{AMUSE}: Emotional Speech-driven {3D} Body Animation via Disentangled Latent Diffusion}. In \bibinfo{booktitle}{\emph{Proceedings of the IEEE/CVF Conference on Computer Vision and Pattern Recognition (CVPR)}}. \bibinfo{pages}{1942--1953}.
\newblock
\urldef\tempurl%
\url{https://amuse.is.tue.mpg.de}
\showURL{%
\tempurl}


\bibitem[Cudeiro et~al\mbox{.}(2019)]%
        {cudeiro2019voca}
\bibfield{author}{\bibinfo{person}{Daniel Cudeiro}, \bibinfo{person}{Timo Bolkart}, \bibinfo{person}{Cassidy Laidlaw}, \bibinfo{person}{Anurag Ranjan}, {and} \bibinfo{person}{Michael~J Black}.} \bibinfo{year}{2019}\natexlab{}.
\newblock \showarticletitle{Capture, learning, and synthesis of 3D speaking styles}. In \bibinfo{booktitle}{\emph{Proceedings of the IEEE/CVF Conference on Computer Vision and Pattern Recognition}}. \bibinfo{pages}{10101--10111}.
\newblock


\bibitem[Danecek et~al\mbox{.}(2022)]%
        {EMOCA:CVPR:2021}
\bibfield{author}{\bibinfo{person}{Radek Danecek}, \bibinfo{person}{Michael~J. Black}, {and} \bibinfo{person}{Timo Bolkart}.} \bibinfo{year}{2022}\natexlab{}.
\newblock \showarticletitle{{EMOCA}: {E}motion Driven Monocular Face Capture and Animation}. In \bibinfo{booktitle}{\emph{Conference on Computer Vision and Pattern Recognition (CVPR)}}. \bibinfo{pages}{20311--20322}.
\newblock


\bibitem[Daněček et~al\mbox{.}(2023)]%
        {emote}
\bibfield{author}{\bibinfo{person}{Radek Daněček}, \bibinfo{person}{Kiran Chhatre}, \bibinfo{person}{Shashank Tripathi}, \bibinfo{person}{Yandong Wen}, \bibinfo{person}{Michael Black}, {and} \bibinfo{person}{Timo Bolkart}.} \bibinfo{year}{2023}\natexlab{}.
\newblock \showarticletitle{Emotional Speech-Driven Animation with Content-Emotion Disentanglement}. \bibinfo{publisher}{ACM}.
\newblock
\urldef\tempurl%
\url{https://doi.org/10.1145/3610548.3618183}
\showDOI{\tempurl}


\bibitem[Edwards et~al\mbox{.}(2016)]%
        {edwards2016jali}
\bibfield{author}{\bibinfo{person}{Pif Edwards}, \bibinfo{person}{Chris Landreth}, \bibinfo{person}{Eugene Fiume}, {and} \bibinfo{person}{Karan Singh}.} \bibinfo{year}{2016}\natexlab{}.
\newblock \showarticletitle{Jali: an animator-centric viseme model for expressive lip synchronization}.
\newblock \bibinfo{journal}{\emph{ACM Transactions on graphics (TOG)}} \bibinfo{volume}{35}, \bibinfo{number}{4} (\bibinfo{year}{2016}), \bibinfo{pages}{1--11}.
\newblock


\bibitem[Egger et~al\mbox{.}(2020)]%
        {Egger2020}
\bibfield{author}{\bibinfo{person}{Bernhard Egger}, \bibinfo{person}{William A.~P. Smith}, \bibinfo{person}{Ayush Tewari}, \bibinfo{person}{Stefanie Wuhrer}, \bibinfo{person}{Michael Zollhoefer}, \bibinfo{person}{Thabo Beeler}, \bibinfo{person}{Florian Bernard}, \bibinfo{person}{Timo Bolkart}, \bibinfo{person}{Adam Kortylewski}, \bibinfo{person}{Sami Romdhani}, \bibinfo{person}{Christian Theobalt}, \bibinfo{person}{Volker Blanz}, {and} \bibinfo{person}{Thomas Vetter}.} \bibinfo{year}{2020}\natexlab{}.
\newblock \showarticletitle{3D Morphable Face Models—Past, Present, and Future}.
\newblock \bibinfo{journal}{\emph{ACM Trans. Graph.}} \bibinfo{volume}{39}, \bibinfo{number}{5}, Article \bibinfo{articleno}{157} (\bibinfo{date}{jun} \bibinfo{year}{2020}), \bibinfo{numpages}{38}~pages.
\newblock
\showISSN{0730-0301}
\urldef\tempurl%
\url{https://doi.org/10.1145/3395208}
\showDOI{\tempurl}


\bibitem[Falcon and {The PyTorch Lightning team}(2019)]%
        {PT_lightning}
\bibfield{author}{\bibinfo{person}{William Falcon} {and} \bibinfo{person}{{The PyTorch Lightning team}}.} \bibinfo{year}{2019}\natexlab{}.
\newblock \bibinfo{booktitle}{\emph{{PyTorch Lightning}}}.
\newblock
\urldef\tempurl%
\url{https://doi.org/10.5281/zenodo.3828935}
\showDOI{\tempurl}


\bibitem[Fan et~al\mbox{.}(2022)]%
        {fan2022faceformer}
\bibfield{author}{\bibinfo{person}{Yingruo Fan}, \bibinfo{person}{Zhaojiang Lin}, \bibinfo{person}{Jun Saito}, \bibinfo{person}{Wenping Wang}, {and} \bibinfo{person}{Taku Komura}.} \bibinfo{year}{2022}\natexlab{}.
\newblock \showarticletitle{FaceFormer: Speech-Driven 3D Facial Animation with Transformers}. In \bibinfo{booktitle}{\emph{Proceedings of the IEEE/CVF Conference on Computer Vision and Pattern Recognition}}. \bibinfo{pages}{18770--18780}.
\newblock


\bibitem[Fanelli et~al\mbox{.}(2010)]%
        {fanelli20103biwi}
\bibfield{author}{\bibinfo{person}{Gabriele Fanelli}, \bibinfo{person}{Juergen Gall}, \bibinfo{person}{Harald Romsdorfer}, \bibinfo{person}{Thibaut Weise}, {and} \bibinfo{person}{Luc Van~Gool}.} \bibinfo{year}{2010}\natexlab{}.
\newblock \showarticletitle{A 3-d audio-visual corpus of affective communication}.
\newblock \bibinfo{journal}{\emph{IEEE Transactions on Multimedia}} \bibinfo{volume}{12}, \bibinfo{number}{6} (\bibinfo{year}{2010}), \bibinfo{pages}{591--598}.
\newblock


\bibitem[Feng et~al\mbox{.}(2021)]%
        {DECA:Siggraph2021}
\bibfield{author}{\bibinfo{person}{Yao Feng}, \bibinfo{person}{Haiwen Feng}, \bibinfo{person}{Michael~J. Black}, {and} \bibinfo{person}{Timo Bolkart}.} \bibinfo{year}{2021}\natexlab{}.
\newblock \showarticletitle{Learning an Animatable Detailed {3D} Face Model from In-The-Wild Images}.
\newblock \bibinfo{journal}{\emph{ACM Transactions on Graphics, (Proc. SIGGRAPH)}} \bibinfo{volume}{40}, \bibinfo{number}{8}.
\newblock
\urldef\tempurl%
\url{https://doi.org/10.1145/3450626.3459936}
\showURL{%
\tempurl}


\bibitem[Giebenhain et~al\mbox{.}(2023)]%
        {giebenhain2023mononphm}
\bibfield{author}{\bibinfo{person}{Simon Giebenhain}, \bibinfo{person}{Tobias Kirschstein}, \bibinfo{person}{Markos Georgopoulos}, \bibinfo{person}{Martin R{\"u}nz}, \bibinfo{person}{Lourdes Agapito}, {and} \bibinfo{person}{Matthias Nie{\ss}ner}.} \bibinfo{year}{2023}\natexlab{}.
\newblock \showarticletitle{MonoNPHM: Dynamic Head Reconstruction from Monocular Videos}.
\newblock \bibinfo{journal}{\emph{arXiv preprint arXiv:2312.06740}} (\bibinfo{year}{2023}).
\newblock


\bibitem[Giebenhain et~al\mbox{.}(2024)]%
        {Giebenhain_2024_CVPR}
\bibfield{author}{\bibinfo{person}{Simon Giebenhain}, \bibinfo{person}{Tobias Kirschstein}, \bibinfo{person}{Markos Georgopoulos}, \bibinfo{person}{Martin R\"unz}, \bibinfo{person}{Lourdes Agapito}, {and} \bibinfo{person}{Matthias Nie{\ss}ner}.} \bibinfo{year}{2024}\natexlab{}.
\newblock \showarticletitle{MonoNPHM: Dynamic Head Reconstruction from Monocular Videos}. In \bibinfo{booktitle}{\emph{Proceedings of the IEEE/CVF Conference on Computer Vision and Pattern Recognition (CVPR)}}. \bibinfo{pages}{10747--10758}.
\newblock


\bibitem[Haque and Yumak(2023)]%
        {FaceXHuBERT_Haque_ICMI23}
\bibfield{author}{\bibinfo{person}{Kazi~Injamamul Haque} {and} \bibinfo{person}{Zerrin Yumak}.} \bibinfo{year}{2023}\natexlab{}.
\newblock \showarticletitle{FaceXHuBERT: Text-less Speech-driven E(X)pressive 3D Facial Animation Synthesis Using Self-Supervised Speech Representation Learning}. In \bibinfo{booktitle}{\emph{INTERNATIONAL CONFERENCE ON MULTIMODAL INTERACTION (ICMI ’23)}} (Paris, France). \bibinfo{publisher}{ACM}, \bibinfo{address}{New York, NY, USA}, \bibinfo{numpages}{10}~pages.
\newblock
\urldef\tempurl%
\url{https://doi.org/10.1145/3577190.3614157}
\showDOI{\tempurl}


\bibitem[Ho et~al\mbox{.}(2020a)]%
        {ddpm}
\bibfield{author}{\bibinfo{person}{Jonathan Ho}, \bibinfo{person}{Ajay Jain}, {and} \bibinfo{person}{Pieter Abbeel}.} \bibinfo{year}{2020}\natexlab{a}.
\newblock \showarticletitle{Denoising diffusion probabilistic models}. In \bibinfo{booktitle}{\emph{Proceedings of the 34th International Conference on Neural Information Processing Systems}} (Vancouver, BC, Canada) \emph{(\bibinfo{series}{NIPS '20})}. \bibinfo{publisher}{Curran Associates Inc.}, \bibinfo{address}{Red Hook, NY, USA}, Article \bibinfo{articleno}{574}, \bibinfo{numpages}{12}~pages.
\newblock
\showISBNx{9781713829546}


\bibitem[Ho et~al\mbox{.}(2020b)]%
        {ho2020denoising}
\bibfield{author}{\bibinfo{person}{Jonathan Ho}, \bibinfo{person}{Ajay Jain}, {and} \bibinfo{person}{Pieter Abbeel}.} \bibinfo{year}{2020}\natexlab{b}.
\newblock \showarticletitle{Denoising diffusion probabilistic models}.
\newblock \bibinfo{journal}{\emph{Advances in Neural Information Processing Systems}}  \bibinfo{volume}{33} (\bibinfo{year}{2020}), \bibinfo{pages}{6840--6851}.
\newblock


\bibitem[Hsu et~al\mbox{.}(2021)]%
        {hubert}
\bibfield{author}{\bibinfo{person}{Wei-Ning Hsu}, \bibinfo{person}{Benjamin Bolte}, \bibinfo{person}{Yao-Hung~Hubert Tsai}, \bibinfo{person}{Kushal Lakhotia}, \bibinfo{person}{Ruslan Salakhutdinov}, {and} \bibinfo{person}{Abdelrahman Mohamed}.} \bibinfo{year}{2021}\natexlab{}.
\newblock \bibinfo{title}{HuBERT: Self-Supervised Speech Representation Learning by Masked Prediction of Hidden Units}.
\newblock
\newblock
\urldef\tempurl%
\url{https://doi.org/10.48550/ARXIV.2106.07447}
\showDOI{\tempurl}


\bibitem[JALI(2023)]%
        {JALI}
JALI \bibinfo{year}{2023}\natexlab{}.
\newblock \bibinfo{title}{JALI Research}.
\newblock
\newblock
\newblock
\shownote{\url{https://jaliresearch.com/}}.


\bibitem[Ji et~al\mbox{.}(2022)]%
        {Xinya2022eamm}
\bibfield{author}{\bibinfo{person}{Xinya Ji}, \bibinfo{person}{Hang Zhou}, \bibinfo{person}{Kaisiyuan Wang}, \bibinfo{person}{Qianyi Wu}, \bibinfo{person}{Wayne Wu}, \bibinfo{person}{Feng Xu}, {and} \bibinfo{person}{Xun Cao}.} \bibinfo{year}{2022}\natexlab{}.
\newblock \showarticletitle{EAMM: One-Shot Emotional Talking Face via Audio-Based Emotion-Aware Motion Model}. In \bibinfo{booktitle}{\emph{ACM SIGGRAPH 2022 Conference Proceedings}} (Vancouver, BC, Canada) \emph{(\bibinfo{series}{SIGGRAPH '22})}. \bibinfo{publisher}{Association for Computing Machinery}, \bibinfo{address}{New York, NY, USA}, Article \bibinfo{articleno}{61}, \bibinfo{numpages}{10}~pages.
\newblock
\showISBNx{9781450393379}
\urldef\tempurl%
\url{https://doi.org/10.1145/3528233.3530745}
\showDOI{\tempurl}


\bibitem[Karras et~al\mbox{.}(2017)]%
        {karras2017audio}
\bibfield{author}{\bibinfo{person}{Tero Karras}, \bibinfo{person}{Timo Aila}, \bibinfo{person}{Samuli Laine}, \bibinfo{person}{Antti Herva}, {and} \bibinfo{person}{Jaakko Lehtinen}.} \bibinfo{year}{2017}\natexlab{}.
\newblock \showarticletitle{Audio-driven facial animation by joint end-to-end learning of pose and emotion}.
\newblock \bibinfo{journal}{\emph{ACM Transactions on Graphics (TOG)}} \bibinfo{volume}{36}, \bibinfo{number}{4} (\bibinfo{year}{2017}), \bibinfo{pages}{1--12}.
\newblock


\bibitem[Li et~al\mbox{.}(2021)]%
        {Jing2021}
\bibfield{author}{\bibinfo{person}{Jing Li}, \bibinfo{person}{Di Kang}, \bibinfo{person}{Wenjie Pei}, \bibinfo{person}{Xuefei Zhe}, \bibinfo{person}{Ying Zhang}, \bibinfo{person}{Zhenyu He}, {and} \bibinfo{person}{Linchao Bao}.} \bibinfo{year}{2021}\natexlab{}.
\newblock \showarticletitle{Audio2Gestures: Generating Diverse Gestures from Speech Audio with Conditional Variational Autoencoders}. In \bibinfo{booktitle}{\emph{2021 IEEE/CVF International Conference on Computer Vision (ICCV)}}. \bibinfo{pages}{11273--11282}.
\newblock
\urldef\tempurl%
\url{https://doi.org/10.1109/ICCV48922.2021.01110}
\showDOI{\tempurl}


\bibitem[Li et~al\mbox{.}(2017a)]%
        {FLAME}
\bibfield{author}{\bibinfo{person}{Tianye Li}, \bibinfo{person}{Timo Bolkart}, \bibinfo{person}{Michael~J. Black}, \bibinfo{person}{Hao Li}, {and} \bibinfo{person}{Javier Romero}.} \bibinfo{year}{2017}\natexlab{a}.
\newblock \showarticletitle{Learning a model of facial shape and expression from 4D scans}.
\newblock \bibinfo{journal}{\emph{ACM Trans. Graph.}} \bibinfo{volume}{36}, \bibinfo{number}{6}, Article \bibinfo{articleno}{194} (\bibinfo{date}{nov} \bibinfo{year}{2017}), \bibinfo{numpages}{17}~pages.
\newblock
\showISSN{0730-0301}
\urldef\tempurl%
\url{https://doi.org/10.1145/3130800.3130813}
\showDOI{\tempurl}


\bibitem[Li et~al\mbox{.}(2017b)]%
        {li2017flame}
\bibfield{author}{\bibinfo{person}{Tianye Li}, \bibinfo{person}{Timo Bolkart}, \bibinfo{person}{Michael~J Black}, \bibinfo{person}{Hao Li}, {and} \bibinfo{person}{Javier Romero}.} \bibinfo{year}{2017}\natexlab{b}.
\newblock \showarticletitle{Learning a model of facial shape and expression from 4D scans.}
\newblock \bibinfo{journal}{\emph{ACM Trans. Graph.}} \bibinfo{volume}{36}, \bibinfo{number}{6} (\bibinfo{year}{2017}), \bibinfo{pages}{194--1}.
\newblock


\bibitem[Liu et~al\mbox{.}(2024)]%
        {emage}
\bibfield{author}{\bibinfo{person}{Haiyang Liu}, \bibinfo{person}{Zihao Zhu}, \bibinfo{person}{Giorgio Becherini}, \bibinfo{person}{Yichen Peng}, \bibinfo{person}{Mingyang Su}, \bibinfo{person}{You Zhou}, \bibinfo{person}{Xuefei Zhe}, \bibinfo{person}{Naoya Iwamoto}, \bibinfo{person}{Bo Zheng}, {and} \bibinfo{person}{Michael~J. Black}.} \bibinfo{year}{2024}\natexlab{}.
\newblock \showarticletitle{{EMAGE}: Towards Unified Holistic Co-Speech Gesture Generation via Expressive Masked Audio Gesture Modeling}. In \bibinfo{booktitle}{\emph{IEEE/CVF Conf.~on Computer Vision and Pattern Recognition (CVPR)}}.
\newblock


\bibitem[Livingstone and Russo(2018)]%
        {livingstone2018ryerson}
\bibfield{author}{\bibinfo{person}{Steven~R Livingstone} {and} \bibinfo{person}{Frank~A Russo}.} \bibinfo{year}{2018}\natexlab{}.
\newblock \showarticletitle{The Ryerson Audio-Visual Database of Emotional Speech and Song (RAVDESS): A dynamic, multimodal set of facial and vocal expressions in North American English}.
\newblock \bibinfo{journal}{\emph{PloS one}} \bibinfo{volume}{13}, \bibinfo{number}{5} (\bibinfo{year}{2018}), \bibinfo{pages}{e0196391}.
\newblock


\bibitem[Lombardi et~al\mbox{.}(2018)]%
        {lombardi2018deep}
\bibfield{author}{\bibinfo{person}{Stephen Lombardi}, \bibinfo{person}{Jason Saragih}, \bibinfo{person}{Tomas Simon}, {and} \bibinfo{person}{Yaser Sheikh}.} \bibinfo{year}{2018}\natexlab{}.
\newblock \showarticletitle{Deep appearance models for face rendering}.
\newblock \bibinfo{journal}{\emph{ACM Transactions on Graphics (ToG)}} \bibinfo{volume}{37}, \bibinfo{number}{4} (\bibinfo{year}{2018}), \bibinfo{pages}{1--13}.
\newblock


\bibitem[Ma et~al\mbox{.}(2024)]%
        {ma2024diffspeaker}
\bibfield{author}{\bibinfo{person}{Zhiyuan Ma}, \bibinfo{person}{Xiangyu Zhu}, \bibinfo{person}{Guojun Qi}, \bibinfo{person}{Chen Qian}, \bibinfo{person}{Zhaoxiang Zhang}, {and} \bibinfo{person}{Zhen Lei}.} \bibinfo{year}{2024}\natexlab{}.
\newblock \showarticletitle{DiffSpeaker: Speech-Driven 3D Facial Animation with Diffusion Transformer}.
\newblock \bibinfo{journal}{\emph{arXiv preprint arXiv:2402.05712}} (\bibinfo{year}{2024}).
\newblock


\bibitem[Ng et~al\mbox{.}(2022)]%
        {ng2022learning2listen}
\bibfield{author}{\bibinfo{person}{Evonne Ng}, \bibinfo{person}{Hanbyul Joo}, \bibinfo{person}{Liwen Hu}, \bibinfo{person}{Hao Li}, \bibinfo{person}{}, \bibinfo{person}{Trevor Darrell}, \bibinfo{person}{Angjoo Kanazawa}, {and} \bibinfo{person}{Shiry Ginosar}.} \bibinfo{year}{2022}\natexlab{}.
\newblock \showarticletitle{Learning to Listen: Modeling Non-Deterministic Dyadic Facial Motion}.
\newblock \bibinfo{journal}{\emph{Proceedings of the IEEE/CVF Conference on Computer Vision and Pattern Recognition}} (\bibinfo{year}{2022}).
\newblock


\bibitem[Ng et~al\mbox{.}(2024)]%
        {ng2024audio2photoreal}
\bibfield{author}{\bibinfo{person}{Evonne Ng}, \bibinfo{person}{Javier Romero}, \bibinfo{person}{Timur Bagautdinov}, \bibinfo{person}{Shaojie Bai}, \bibinfo{person}{Trevor Darrell}, \bibinfo{person}{Angjoo Kanazawa}, {and} \bibinfo{person}{Alexander Richard}.} \bibinfo{year}{2024}\natexlab{}.
\newblock \showarticletitle{From Audio to Photoreal Embodiment: Synthesizing Humans in Conversations}. In \bibinfo{booktitle}{\emph{IEEE Conference on Computer Vision and Pattern Recognition}}.
\newblock


\bibitem[Park and Cho(2023)]%
        {park2023said}
\bibfield{author}{\bibinfo{person}{Inkyu Park} {and} \bibinfo{person}{Jaewoong Cho}.} \bibinfo{year}{2023}\natexlab{}.
\newblock \bibinfo{title}{SAiD: Speech-driven Blendshape Facial Animation with Diffusion}.
\newblock
\newblock
\showeprint[arxiv]{2401.08655}~[cs.CV]


\bibitem[Park et~al\mbox{.}(2023)]%
        {park2023df}
\bibfield{author}{\bibinfo{person}{Se~Jin Park}, \bibinfo{person}{Joanna Hong}, \bibinfo{person}{Minsu Kim}, {and} \bibinfo{person}{Yong~Man Ro}.} \bibinfo{year}{2023}\natexlab{}.
\newblock \showarticletitle{DF-3DFace: One-to-Many Speech Synchronized 3D Face Animation with Diffusion}.
\newblock \bibinfo{journal}{\emph{arXiv preprint arXiv:2310.05934}} (\bibinfo{year}{2023}).
\newblock


\bibitem[Peng et~al\mbox{.}(2023)]%
        {peng2023emotalk}
\bibfield{author}{\bibinfo{person}{Ziqiao Peng}, \bibinfo{person}{Haoyu Wu}, \bibinfo{person}{Zhenbo Song}, \bibinfo{person}{Hao Xu}, \bibinfo{person}{Xiangyu Zhu}, \bibinfo{person}{Hongyan Liu}, \bibinfo{person}{Jun He}, {and} \bibinfo{person}{Zhaoxin Fan}.} \bibinfo{year}{2023}\natexlab{}.
\newblock \showarticletitle{EmoTalk: Speech-Driven Emotional Disentanglement for 3D Face Animation}. In \bibinfo{booktitle}{\emph{Proceedings of the IEEE/CVF international conference on computer vision}}.
\newblock


\bibitem[Prolific(2023)]%
        {Prolific}
Prolific \bibinfo{year}{2023}\natexlab{}.
\newblock \bibinfo{title}{Prolific}.
\newblock
\newblock
\newblock
\shownote{\url{https://www.prolific.co}}.


\bibitem[Qualtrics(2023)]%
        {qualtrics}
Qualtrics \bibinfo{year}{2023}\natexlab{}.
\newblock \bibinfo{title}{Qualtrics}.
\newblock
\newblock
\newblock
\shownote{\url{https://www.qualtrics.com}}.


\bibitem[Radford et~al\mbox{.}(2021)]%
        {CLIP}
\bibfield{author}{\bibinfo{person}{Alec Radford}, \bibinfo{person}{Jong~Wook Kim}, \bibinfo{person}{Chris Hallacy}, \bibinfo{person}{Aditya Ramesh}, \bibinfo{person}{Gabriel Goh}, \bibinfo{person}{Sandhini Agarwal}, \bibinfo{person}{Girish Sastry}, \bibinfo{person}{Amanda Askell}, \bibinfo{person}{Pamela Mishkin}, \bibinfo{person}{Jack Clark}, \bibinfo{person}{Gretchen Krueger}, {and} \bibinfo{person}{Ilya Sutskever}.} \bibinfo{year}{2021}\natexlab{}.
\newblock \showarticletitle{Learning Transferable Visual Models From Natural Language Supervision}. In \bibinfo{booktitle}{\emph{Proceedings of the 38th International Conference on Machine Learning}} \emph{(\bibinfo{series}{Proceedings of Machine Learning Research}, Vol.~\bibinfo{volume}{139})}, \bibfield{editor}{\bibinfo{person}{Marina Meila} {and} \bibinfo{person}{Tong Zhang}} (Eds.). \bibinfo{publisher}{PMLR}, \bibinfo{pages}{8748--8763}.
\newblock
\urldef\tempurl%
\url{https://proceedings.mlr.press/v139/radford21a.html}
\showURL{%
\tempurl}


\bibitem[Ren et~al\mbox{.}(2023)]%
        {ren2023diffusion}
\bibfield{author}{\bibinfo{person}{Zhiyuan Ren}, \bibinfo{person}{Zhihong Pan}, \bibinfo{person}{Xin Zhou}, {and} \bibinfo{person}{Le Kang}.} \bibinfo{year}{2023}\natexlab{}.
\newblock \showarticletitle{Diffusion motion: Generate text-guided 3d human motion by diffusion model}. In \bibinfo{booktitle}{\emph{ICASSP 2023-2023 IEEE International Conference on Acoustics, Speech and Signal Processing (ICASSP)}}. IEEE, \bibinfo{pages}{1--5}.
\newblock


\bibitem[Richard et~al\mbox{.}(2021a)]%
        {richard2021audio}
\bibfield{author}{\bibinfo{person}{Alexander Richard}, \bibinfo{person}{Colin Lea}, \bibinfo{person}{Shugao Ma}, \bibinfo{person}{Jurgen Gall}, \bibinfo{person}{Fernando De~la Torre}, {and} \bibinfo{person}{Yaser Sheikh}.} \bibinfo{year}{2021}\natexlab{a}.
\newblock \showarticletitle{Audio-and gaze-driven facial animation of codec avatars}. In \bibinfo{booktitle}{\emph{Proceedings of the IEEE/CVF winter conference on applications of computer vision}}. \bibinfo{pages}{41--50}.
\newblock


\bibitem[Richard et~al\mbox{.}(2021b)]%
        {richard2021meshtalk}
\bibfield{author}{\bibinfo{person}{Alexander Richard}, \bibinfo{person}{Michael Zollh{\"o}fer}, \bibinfo{person}{Yandong Wen}, \bibinfo{person}{Fernando De~la Torre}, {and} \bibinfo{person}{Yaser Sheikh}.} \bibinfo{year}{2021}\natexlab{b}.
\newblock \showarticletitle{Meshtalk: 3d face animation from speech using cross-modality disentanglement}. In \bibinfo{booktitle}{\emph{Proceedings of the IEEE/CVF International Conference on Computer Vision}}. \bibinfo{pages}{1173--1182}.
\newblock


\bibitem[Stan et~al\mbox{.}(2023)]%
        {FaceDiffuser_Stan_MIG2023}
\bibfield{author}{\bibinfo{person}{Stefan Stan}, \bibinfo{person}{Kazi~Injamamul Haque}, {and} \bibinfo{person}{Zerrin Yumak}.} \bibinfo{year}{2023}\natexlab{}.
\newblock \showarticletitle{FaceDiffuser: Speech-Driven 3D Facial Animation Synthesis Using Diffusion}. In \bibinfo{booktitle}{\emph{ACM SIGGRAPH Conference on Motion, Interaction and Games (MIG '23), November 15--17, 2023, Rennes, France}} (Rennes, France). \bibinfo{publisher}{ACM}, \bibinfo{address}{New York, NY, USA}, \bibinfo{numpages}{11}~pages.
\newblock
\urldef\tempurl%
\url{https://doi.org/10.1145/3623264.3624447}
\showDOI{\tempurl}


\bibitem[Stypu{\l}kowski et~al\mbox{.}(2024)]%
        {Stypulkowski_2024_WACV}
\bibfield{author}{\bibinfo{person}{Micha{\l} Stypu{\l}kowski}, \bibinfo{person}{Konstantinos Vougioukas}, \bibinfo{person}{Sen He}, \bibinfo{person}{Maciej Zi\k{e}ba}, \bibinfo{person}{Stavros Petridis}, {and} \bibinfo{person}{Maja Pantic}.} \bibinfo{year}{2024}\natexlab{}.
\newblock \showarticletitle{Diffused Heads: Diffusion Models Beat GANs on Talking-Face Generation}. In \bibinfo{booktitle}{\emph{Proceedings of the IEEE/CVF Winter Conference on Applications of Computer Vision (WACV)}}. \bibinfo{pages}{5091--5100}.
\newblock


\bibitem[Sun et~al\mbox{.}(2024)]%
        {sun2023diffposetalk}
\bibfield{author}{\bibinfo{person}{Zhiyao Sun}, \bibinfo{person}{Tian Lv}, \bibinfo{person}{Sheng Ye}, \bibinfo{person}{Matthieu Lin}, \bibinfo{person}{Jenny Sheng}, \bibinfo{person}{Yu-Hui Wen}, \bibinfo{person}{Minjing Yu}, {and} \bibinfo{person}{Yong-Jin Liu}.} \bibinfo{year}{2024}\natexlab{}.
\newblock \showarticletitle{DiffPoseTalk: Speech-Driven Stylistic 3D Facial Animation and Head Pose Generation via Diffusion Models}.
\newblock \bibinfo{journal}{\emph{ACM Trans. Graph.}} \bibinfo{volume}{43}, \bibinfo{number}{4}, Article \bibinfo{articleno}{46} (\bibinfo{date}{jul} \bibinfo{year}{2024}), \bibinfo{numpages}{9}~pages.
\newblock
\showISSN{0730-0301}
\urldef\tempurl%
\url{https://doi.org/10.1145/3658221}
\showDOI{\tempurl}


\bibitem[Taylor et~al\mbox{.}(2017)]%
        {taylor2017deep}
\bibfield{author}{\bibinfo{person}{Sarah Taylor}, \bibinfo{person}{Taehwan Kim}, \bibinfo{person}{Yisong Yue}, \bibinfo{person}{Moshe Mahler}, \bibinfo{person}{James Krahe}, \bibinfo{person}{Anastasio~Garcia Rodriguez}, \bibinfo{person}{Jessica Hodgins}, {and} \bibinfo{person}{Iain Matthews}.} \bibinfo{year}{2017}\natexlab{}.
\newblock \showarticletitle{A deep learning approach for generalized speech animation}.
\newblock \bibinfo{journal}{\emph{ACM Transactions on Graphics (TOG)}} \bibinfo{volume}{36}, \bibinfo{number}{4} (\bibinfo{year}{2017}), \bibinfo{pages}{1--11}.
\newblock


\bibitem[Taylor et~al\mbox{.}(2012)]%
        {Taylor2012}
\bibfield{author}{\bibinfo{person}{Sarah~L. Taylor}, \bibinfo{person}{Moshe Mahler}, \bibinfo{person}{Barry-John Theobald}, {and} \bibinfo{person}{Iain Matthews}.} \bibinfo{year}{2012}\natexlab{}.
\newblock \showarticletitle{Dynamic Units of Visual Speech}. In \bibinfo{booktitle}{\emph{Proceedings of the ACM SIGGRAPH/Eurographics Symposium on Computer Animation}} (Lausanne, Switzerland) \emph{(\bibinfo{series}{SCA '12})}. \bibinfo{publisher}{Eurographics Association}, \bibinfo{address}{Goslar, DEU}, \bibinfo{pages}{275–284}.
\newblock
\showISBNx{9783905674378}


\bibitem[Tevet et~al\mbox{.}(2022)]%
        {tevet2022human}
\bibfield{author}{\bibinfo{person}{Guy Tevet}, \bibinfo{person}{Sigal Raab}, \bibinfo{person}{Brian Gordon}, \bibinfo{person}{Yonatan Shafir}, \bibinfo{person}{Daniel Cohen-Or}, {and} \bibinfo{person}{Amit~H Bermano}.} \bibinfo{year}{2022}\natexlab{}.
\newblock \showarticletitle{Human motion diffusion model}.
\newblock \bibinfo{journal}{\emph{arXiv preprint arXiv:2209.14916}} (\bibinfo{year}{2022}).
\newblock


\bibitem[Thambiraja et~al\mbox{.}(2023a)]%
        {3diface}
\bibfield{author}{\bibinfo{person}{Balamurugan Thambiraja}, \bibinfo{person}{Sadegh Aliakbarian}, \bibinfo{person}{Darren Cosker}, {and} \bibinfo{person}{Justus Thies}.} \bibinfo{year}{2023}\natexlab{a}.
\newblock \bibinfo{title}{3DiFACE: Diffusion-based Speech-driven 3D Facial Animation and Editing}.
\newblock
\newblock
\showeprint[arxiv]{2312.00870}~[cs.CV]
\urldef\tempurl%
\url{https://arxiv.org/abs/2312.00870}
\showURL{%
\tempurl}


\bibitem[Thambiraja et~al\mbox{.}(2023b)]%
        {thambiraja20233diface}
\bibfield{author}{\bibinfo{person}{Balamurugan Thambiraja}, \bibinfo{person}{Sadegh Aliakbarian}, \bibinfo{person}{Darren Cosker}, {and} \bibinfo{person}{Justus Thies}.} \bibinfo{year}{2023}\natexlab{b}.
\newblock \showarticletitle{3DiFACE: Diffusion-based Speech-driven 3D Facial Animation and Editing}.
\newblock \bibinfo{journal}{\emph{arXiv preprint arXiv:2312.00870}} (\bibinfo{year}{2023}).
\newblock


\bibitem[Thambiraja et~al\mbox{.}(2023c)]%
        {thambiraja2023imitator}
\bibfield{author}{\bibinfo{person}{Balamurugan Thambiraja}, \bibinfo{person}{Ikhsanul Habibie}, \bibinfo{person}{Sadegh Aliakbarian}, \bibinfo{person}{Darren Cosker}, \bibinfo{person}{Christian Theobalt}, {and} \bibinfo{person}{Justus Thies}.} \bibinfo{year}{2023}\natexlab{c}.
\newblock \showarticletitle{Imitator: Personalized speech-driven 3d facial animation}. In \bibinfo{booktitle}{\emph{Proceedings of the IEEE/CVF International Conference on Computer Vision}}. \bibinfo{pages}{20621--20631}.
\newblock


\bibitem[Van Den~Oord et~al\mbox{.}(2017)]%
        {van2017neural}
\bibfield{author}{\bibinfo{person}{Aaron Van Den~Oord}, \bibinfo{person}{Oriol Vinyals}, {et~al\mbox{.}}} \bibinfo{year}{2017}\natexlab{}.
\newblock \showarticletitle{Neural discrete representation learning}.
\newblock \bibinfo{journal}{\emph{Advances in neural information processing systems}}  \bibinfo{volume}{30} (\bibinfo{year}{2017}).
\newblock


\bibitem[van~den Oord et~al\mbox{.}(2017)]%
        {VQVAE}
\bibfield{author}{\bibinfo{person}{Aaron van~den Oord}, \bibinfo{person}{Oriol Vinyals}, {and} \bibinfo{person}{Koray Kavukcuoglu}.} \bibinfo{year}{2017}\natexlab{}.
\newblock \showarticletitle{Neural discrete representation learning}. In \bibinfo{booktitle}{\emph{Proceedings of the 31st International Conference on Neural Information Processing Systems}} (Long Beach, California, USA) \emph{(\bibinfo{series}{NIPS'17})}. \bibinfo{publisher}{Curran Associates Inc.}, \bibinfo{address}{Red Hook, NY, USA}, \bibinfo{pages}{6309–6318}.
\newblock
\showISBNx{9781510860964}


\bibitem[Wang et~al\mbox{.}(2020)]%
        {kaisiyuan2020mead}
\bibfield{author}{\bibinfo{person}{Kaisiyuan Wang}, \bibinfo{person}{Qianyi Wu}, \bibinfo{person}{Linsen Song}, \bibinfo{person}{Zhuoqian Yang}, \bibinfo{person}{Wayne Wu}, \bibinfo{person}{Chen Qian}, \bibinfo{person}{Ran He}, \bibinfo{person}{Yu Qiao}, {and} \bibinfo{person}{Chen~Change Loy}.} \bibinfo{year}{2020}\natexlab{}.
\newblock \showarticletitle{MEAD: A Large-scale Audio-visual Dataset for Emotional Talking-face Generation}. In \bibinfo{booktitle}{\emph{ECCV}}.
\newblock


\bibitem[Wuu et~al\mbox{.}(2022)]%
        {wuu2022multiface}
\bibfield{author}{\bibinfo{person}{Cheng-hsin Wuu}, \bibinfo{person}{Ningyuan Zheng}, \bibinfo{person}{Scott Ardisson}, \bibinfo{person}{Rohan Bali}, \bibinfo{person}{Danielle Belko}, \bibinfo{person}{Eric Brockmeyer}, \bibinfo{person}{Lucas Evans}, \bibinfo{person}{Timothy Godisart}, \bibinfo{person}{Hyowon Ha}, \bibinfo{person}{Alexander Hypes}, {et~al\mbox{.}}} \bibinfo{year}{2022}\natexlab{}.
\newblock \showarticletitle{Multiface: A dataset for neural face rendering}.
\newblock \bibinfo{journal}{\emph{arXiv preprint arXiv:2207.11243}} (\bibinfo{year}{2022}).
\newblock


\bibitem[Xing et~al\mbox{.}(2023)]%
        {xing2023codetalker}
\bibfield{author}{\bibinfo{person}{Jinbo Xing}, \bibinfo{person}{Menghan Xia}, \bibinfo{person}{Yuechen Zhang}, \bibinfo{person}{Xiaodong Cun}, \bibinfo{person}{Jue Wang}, {and} \bibinfo{person}{Tien-Tsin Wong}.} \bibinfo{year}{2023}\natexlab{}.
\newblock \showarticletitle{CodeTalker: Speech-Driven 3D Facial Animation with Discrete Motion Prior}.
\newblock \bibinfo{journal}{\emph{arXiv preprint arXiv:2301.02379}} (\bibinfo{year}{2023}).
\newblock


\bibitem[Yang et~al\mbox{.}(2024)]%
        {yang2024probabilistic}
\bibfield{author}{\bibinfo{person}{Karren~D Yang}, \bibinfo{person}{Anurag Ranjan}, \bibinfo{person}{Jen-Hao~Rick Chang}, \bibinfo{person}{Raviteja Vemulapalli}, {and} \bibinfo{person}{Oncel Tuzel}.} \bibinfo{year}{2024}\natexlab{}.
\newblock \showarticletitle{Probabilistic Speech-Driven 3D Facial Motion Synthesis: New Benchmarks Methods and Applications}. In \bibinfo{booktitle}{\emph{Proceedings of the IEEE/CVF Conference on Computer Vision and Pattern Recognition}}. \bibinfo{pages}{27294--27303}.
\newblock


\bibitem[Yi et~al\mbox{.}(2023)]%
        {yi2023generating}
\bibfield{author}{\bibinfo{person}{Hongwei Yi}, \bibinfo{person}{Hualin Liang}, \bibinfo{person}{Yifei Liu}, \bibinfo{person}{Qiong Cao}, \bibinfo{person}{Yandong Wen}, \bibinfo{person}{Timo Bolkart}, \bibinfo{person}{Dacheng Tao}, {and} \bibinfo{person}{Michael~J Black}.} \bibinfo{year}{2023}\natexlab{}.
\newblock \showarticletitle{Generating Holistic 3D Human Motion from Speech}. In \bibinfo{booktitle}{\emph{IEEE Conference on Computer Vision and Pattern Recognition (CVPR)}}. \bibinfo{pages}{469--480}.
\newblock


\bibitem[Zhang et~al\mbox{.}(2023)]%
        {zhang2023sadtalker}
\bibfield{author}{\bibinfo{person}{Wenxuan Zhang}, \bibinfo{person}{Xiaodong Cun}, \bibinfo{person}{Xuan Wang}, \bibinfo{person}{Yong Zhang}, \bibinfo{person}{Xi Shen}, \bibinfo{person}{Yu Guo}, \bibinfo{person}{Ying Shan}, {and} \bibinfo{person}{Fei Wang}.} \bibinfo{year}{2023}\natexlab{}.
\newblock \showarticletitle{SadTalker: Learning Realistic 3D Motion Coefficients for Stylized Audio-Driven Single Image Talking Face Animation}. In \bibinfo{booktitle}{\emph{Proceedings of the IEEE/CVF Conference on Computer Vision and Pattern Recognition}}. \bibinfo{pages}{8652--8661}.
\newblock


\bibitem[Zhang et~al\mbox{.}(2021)]%
        {zhang2021flow}
\bibfield{author}{\bibinfo{person}{Zhimeng Zhang}, \bibinfo{person}{Lincheng Li}, \bibinfo{person}{Yu Ding}, {and} \bibinfo{person}{Changjie Fan}.} \bibinfo{year}{2021}\natexlab{}.
\newblock \showarticletitle{Flow-guided one-shot talking face generation with a high-resolution audio-visual dataset}. In \bibinfo{booktitle}{\emph{Proceedings of the IEEE/CVF Conference on Computer Vision and Pattern Recognition}}. \bibinfo{pages}{3661--3670}.
\newblock


\bibitem[Zhao et~al\mbox{.}(2024)]%
        {zhao2024media2face}
\bibfield{author}{\bibinfo{person}{Qingcheng Zhao}, \bibinfo{person}{Pengyu Long}, \bibinfo{person}{Qixuan Zhang}, \bibinfo{person}{Dafei Qin}, \bibinfo{person}{Han Liang}, \bibinfo{person}{Longwen Zhang}, \bibinfo{person}{Yingliang Zhang}, \bibinfo{person}{Jingyi Yu}, {and} \bibinfo{person}{Lan Xu}.} \bibinfo{year}{2024}\natexlab{}.
\newblock \showarticletitle{Media2Face: Co-speech Facial Animation Generation With Multi-Modality Guidance}. In \bibinfo{booktitle}{\emph{ACM SIGGRAPH 2024 Conference Papers}} (Denver, CO, USA) \emph{(\bibinfo{series}{SIGGRAPH '24})}. \bibinfo{publisher}{Association for Computing Machinery}, \bibinfo{address}{New York, NY, USA}, Article \bibinfo{articleno}{18}, \bibinfo{numpages}{13}~pages.
\newblock
\showISBNx{9798400705250}
\urldef\tempurl%
\url{https://doi.org/10.1145/3641519.3657413}
\showDOI{\tempurl}


\bibitem[Zhao et~al\mbox{.}(2023)]%
        {zhao2023breathing}
\bibfield{author}{\bibinfo{person}{Wei Zhao}, \bibinfo{person}{Yijun Wang}, \bibinfo{person}{Tianyu He}, \bibinfo{person}{Lianying Yin}, \bibinfo{person}{Jianxin Lin}, {and} \bibinfo{person}{Xin Jin}.} \bibinfo{year}{2023}\natexlab{}.
\newblock \showarticletitle{Breathing Life into Faces: Speech-driven 3D Facial Animation with Natural Head Pose and Detailed Shape}.
\newblock \bibinfo{journal}{\emph{arXiv preprint arXiv:2310.20240}} (\bibinfo{year}{2023}).
\newblock


\bibitem[Zhu et~al\mbox{.}(2024)]%
        {surveybodymotion}
\bibfield{author}{\bibinfo{person}{Wentao Zhu}, \bibinfo{person}{Xiaoxuan Ma}, \bibinfo{person}{Dongwoo Ro}, \bibinfo{person}{Hai Ci}, \bibinfo{person}{Jinlu Zhang}, \bibinfo{person}{Jiaxin Shi}, \bibinfo{person}{Feng Gao}, \bibinfo{person}{Qi Tian}, {and} \bibinfo{person}{Yizhou Wang}.} \bibinfo{year}{2024}\natexlab{}.
\newblock \showarticletitle{Human Motion Generation: A Survey}.
\newblock \bibinfo{journal}{\emph{IEEE Transactions on Pattern Analysis and Machine Intelligence}} \bibinfo{volume}{46}, \bibinfo{number}{4} (\bibinfo{year}{2024}), \bibinfo{pages}{2430--2449}.
\newblock
\urldef\tempurl%
\url{https://doi.org/10.1109/TPAMI.2023.3330935}
\showDOI{\tempurl}


\bibitem[Zielonka et~al\mbox{.}(2022)]%
        {MICA}
\bibfield{author}{\bibinfo{person}{Wojciech Zielonka}, \bibinfo{person}{Timo Bolkart}, {and} \bibinfo{person}{Justus Thies}.} \bibinfo{year}{2022}\natexlab{}.
\newblock \showarticletitle{Towards Metrical Reconstruction of Human Faces}. In \bibinfo{booktitle}{\emph{Computer Vision – ECCV 2022}} \emph{(\bibinfo{series}{Lecture Notes in Computer Science, 13673}, Vol.~\bibinfo{volume}{13})}. \bibinfo{publisher}{Springer}, \bibinfo{address}{Cham}, \bibinfo{pages}{250--269}.
\newblock
\urldef\tempurl%
\url{https://doi.org/10.1007/978-3-031-19778-9_15}
\showDOI{\tempurl}


\end{thebibliography}

\clearpage
\appendix

\section{Supplementary Material}

\subsection{Dataset and Split}
FLAME parameter datasets have been constructed in several studies \cite{ng2022learning2listen, sun2023diffposetalk, park2023df, zhao2023breathing}. As stated in \cite{sun2023diffposetalk}, the use of lower-dimensional 3DMM parameters allows for faster computational speed compared to predicting mesh vertices which are typically much higher dimensional than 3DMM parameters (analogous to blendshapes and morph targets). More recently, EMOTE\cite{emote} reconstructed 3DMEAD (a FLAME parameter dataset) from videos of audio-visual dataset MEAD \cite{kaisiyuan2020mead}. 3DMEAD includes the facial animation data (in terms of reconstructed FLAME 3DMM parameters), audios, emotion annotations and intensity annotations from the original MEAD dataset. The emotion and intensity annotations makes this dataset useful for our work as it is the only large-scale audio-3D dataset with labeled emotion variations that can be accessed and used publicly for research and academic purposes. 

EMOTE \cite{emote} uses all sequences from 32 subjects for training and 7 subjects for validation, without any sequences allocated for evaluation. However, we find this split insufficient for objective model evaluation. In our 2-stage approach, the first stage trains a prior model to learn from the motion data sequences. To maximize data utilization, we adopt the EMOTE split. For the second stage which involves audio-to-motion generation, we propose a novel split inspired by the dataset split of BIWI dataset \cite{fanelli20103biwi} as also used in \cite{fan2022faceformer, FaceXHuBERT_Haque_ICMI23, xing2023codetalker, FaceDiffuser_Stan_MIG2023} that lets us evaluate the results with objective metrics. We divide the sentences of the 32 training subjects into training, validation, and test sets in an 80-10-10 ratio. The training subjects utilized are identical to those in EMOTE, facilitating a meaningful performance comparison during perceptual evaluation. Additionally, given the scale of 3DMEAD, we believe our split provides sufficient data for effective learning and generation. To elaborate further, each subject utters 40 sentences in the neutral expression. We allocate the first 32 sentences for training, the subsequent 4 for validation, and the remaining 4 for evaluation. For each of the 7 emotions, every subject performs 30 sentences across 3 intensity levels. Similarly, we assign the first 24 sentences (including all intensities) for training, the next 3 for validation, and the remaining 3 for evaluation. The split is summarized in Table \ref{tab:3D MEAD split}. When examining the sequence numbers in the table, please be aware that the reconstructed 3DMEAD dataset is missing some sequences from the original 2D MEAD dataset. For the user study, we then utilize sequences from the remaining $(47-32-7=8)$ unseen identities, as well as the unseen audio during training (including the test set in stage 2 and the validation set in stage 1). Specifically, we want to explain our choice of data split for stage 1. Intuitively, the data split for both stages should be the same. However, our experiments show that using less data in stage 1 can significantly affect the performance of the motion prior. Therefore, we adopt the stage 1 split shown in Table \ref{tab:3D MEAD split}.

\begin{table*}[]
    \centering
    \begin{tabular}{>{\centering\arraybackslash}p{3cm}>{\centering\arraybackslash}p{4.8cm}>{\centering\arraybackslash}p{4.8cm}}
    \hline
    \textbf{Dataset} & \textbf{Stage-1 split} & \textbf{Stage-2 split} \\
      &   &   \\
    \hline
    Training set & 
    \begin{tabular}[c]{@{}c@{}}32 identities \\ $40+(30\times7\times3)$ sentences \\ = 21,115 sequences\end{tabular}  & 
    \begin{tabular}[c]{@{}c@{}}32 identities \\ $32+(24\times7\times3)$ sentences \\ = 17,098 sequences\end{tabular} \\
    \hline
    Validation set &
    \begin{tabular}[c]{@{}c@{}}7 identities \\ $40+(30\times7\times3)$ sentences \\ = 4,584 sequences\end{tabular} &
    \begin{tabular}[c]{@{}c@{}}32 identities \\ $4+(3\times7\times3)$ sentences \\ = 2,108 sequences\end{tabular} \\
    \hline
    Test set & \textbf{x} &
    \begin{tabular}[c]{@{}c@{}}32 identities \\ $4+(3\times7\times3)$ sentences \\ = 1,909 sequences\end{tabular} \\
    \hline
    \end{tabular}
    \vspace{1pt}
    \caption{Dataset split of 3DMEAD used in our experiments.}
    \label{tab:3D MEAD split}
\end{table*}

\subsection{VQ-VAE Background}
VQ-VAE \cite{van2017neural} introduces a novel approach by using a discrete codebook to categorically model the input data, rather than using a continuous representation. In a VQ-VAE, the encoder outputs go through a procedure known as vector quantization. Specifically, this involves comparing the encoder output to vectors embedded in a codebook. The codebook comprises discrete vectors, each representing a different category in the data. Initially, the VQ-VAE algorithm selects the nearest vector in the codebook to the encoder output, a process inherently deterministic. However, by incorporating stochastic sampling techniques, the model's behavior can be transformed into non-deterministic. The prior distribution $P(z)$ is a categorical distribution in a VQ-VAE. During the generation process, an embedding vector is sampled from the distribution and decoded to motion. The vector quantization methodology mitigates the posterior collapse often observed in VAEs, where latent variables may be ignored, particularly when coupled with a powerful decoder \cite{van2017neural}. This occurs when the learned latent variables become uninformative due to an overemphasis on minimizing the KL divergence term. The decoder might produce similar outputs, regardless of the latent variable input. Furthermore, the use of a discrete codebook contributes to increased memory efficiency, showcasing its ability to capture crucial features of the input data within a compressed, discrete latent space.  In the context of a VQ-VAE, $z_e(x)$ denotes the output of the encoder, $e_i$ represents the $i$-th codebook vector, and $Q(z|x)$ is the approximate posterior categorical distribution, defined as one-hot vectors:
\begin{equation} \label{equation_vqvae_post}
Q(z = k | x) =
\begin{cases}
1 & \text{for } k = \text{argmin}_{i} \|z_e(x) - e_i\|_2, \\
0 & \text{otherwise.}
\end{cases}
\end{equation}
In the given equation, $k$ is the codebook index corresponding to the embedding $e$ with the minimum distance from $z_e(x)$. The quantization process maps the encoder outputs $z_e(x)$ to the nearest embedding vectors in the codebook, resulting in a quantized vector $z_Q(x)$. It can be formulated as equation \eqref{equation_vqvae_quant}.
\begin{equation} \label{equation_vqvae_quant}
z_Q(x) = e_k,  \text{where } k = \text{argmin}_{i} \|z_e(x) - e_i\|_2
\end{equation}
The total training objective is expressed as:
\begin{equation} \label{equation_vqvae_loss}
\mathcal{L} = \log P(x | z_Q(x)) + \|\text{sg}[z_e(x)] - e\|_2^2 + \beta \|z_e(x) - \text{sg}[e]\|_2^2
\end{equation}
In this function, $\log P(x | z_Q(x))$ denotes the reconstruction loss, while $\|\text{sg}[z_e(x)] - e\|_2^2$ represents the codebook loss, move the codebook embedding vectors $e_i$ towards the encoder outputs. Additionally, $\|z_e(x) - \text{sg}[e]\|_2^2$ represents the commitment loss, ensuring the encoder outputs commit to specific embeddings. Here, $sg$ stands for the stop gradient operator, and $\beta$ is a hyperparameter controlling the weight of the commitment loss.

\subsection{User Study}
We post the survey on Prolific \cite{Prolific} to gather responses. The platform allows researchers to pay a fair amount for participants and provides solutions for addressing privacy and ethical issues. Participants who speak fluent English are recruited through Prolific, providing 40 responses. Additionally, we find 33 participants online who voluntarily took part in the perceptual study in kind. In total, 73 responses are collected, with 4 responses discarded due to failing the attention test, resulting in 69 valid responses. Demographically, 24.64\% of participants are aged 18 to 23, 33.33\% aged 24 to 28, 14.49\% aged 29 to 33, and 27.54\% are above 33 years old. Participants are asked to rate their familiarity with virtual humans, 3D animated films, and video games on a scale of 0 to 7 at the beginning of the user study. Among the 69 participants, the average familiarity scores are 3.90 for virtual humans, 5.22 for 3D animated films, and 5.52 for video games.

All audio samples selected for the user study are between 3 and 9 seconds. Note that since our model and EMOTE only produces output with expression and jaw parameters, we post-process the outputs to add an average shape of specific subjects and transfer them to the vertex space for rendering. This is done to align with the face shapes in the outputs of FaceDiffuser. We apply the same post-processing to the ground truth, using an average shape for a fair comparison.

\begin{figure}[H]
\begin{subfigure}{.5\textwidth}
  \centering
  \includegraphics[scale=0.45]{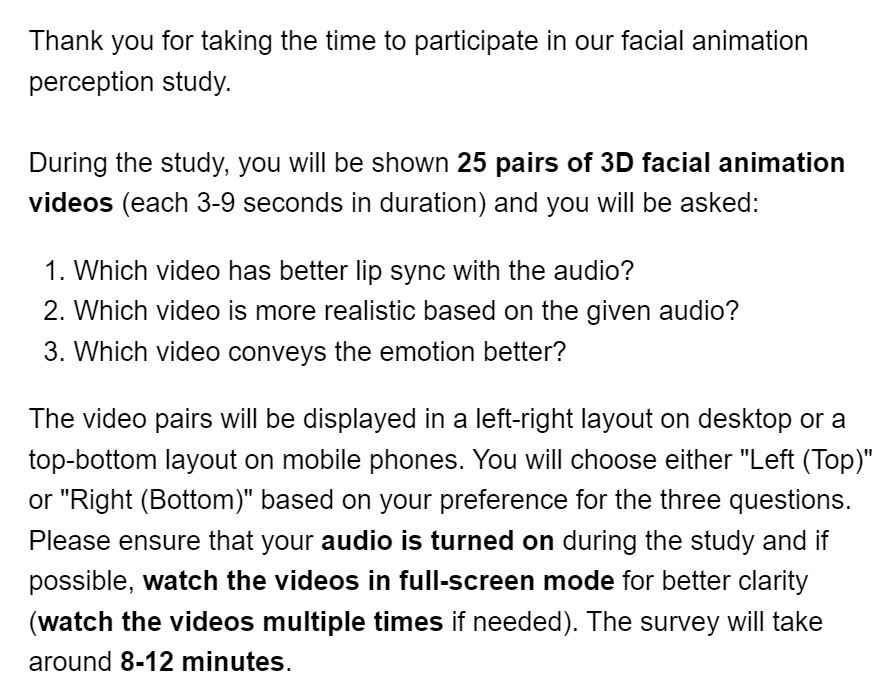}  
  \caption{User study instructions.}
  \label{fig:user_study_deck_instruct}
\end{subfigure}
\begin{subfigure}{.5\textwidth}
  \centering
  \includegraphics[scale=0.45]{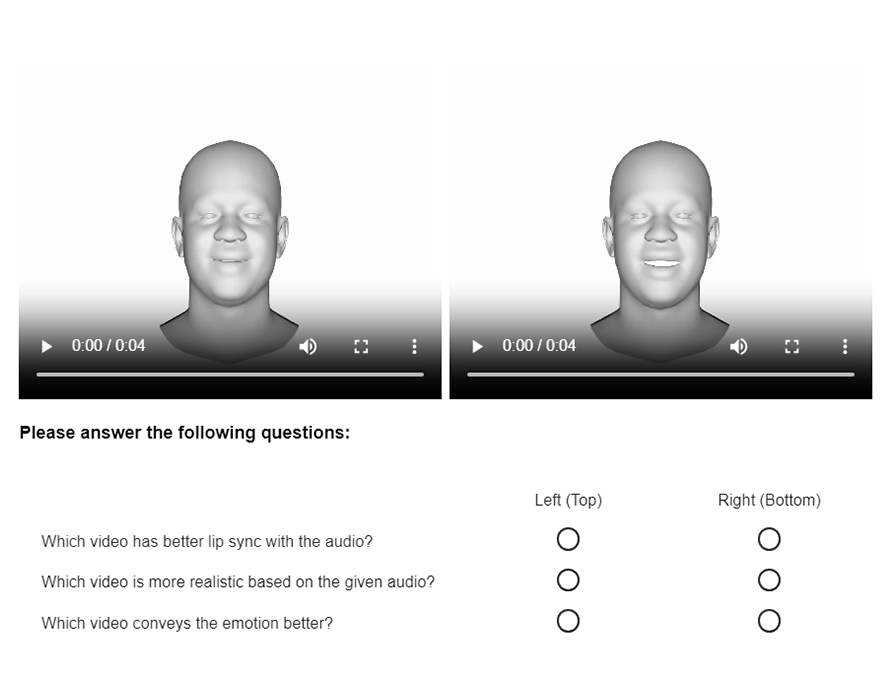}  
  \caption{UI of the perceptual user study.}
  \label{fig:user_study_desk_layout}
\end{subfigure}
\caption{Example screenshots of the user study displayed on desktop environment.}
\label{fig:user_study_desk}
\end{figure}

An attention test is incorporated into the study to filter out unreliable responses, where one video intentionally has significant video and audio out of sync. Failure to correctly identify the higher-quality video in this test leads to early termination of the survey and disregarding the response. Altogether, 25 video pairs (including the attention test) are presented to the participants in random order. For each video pair, two videos are displayed in a left-right layout on desktops or a top-bottom layout on mobile phones. One video showcases the output of our model, while the other showcases either the ground truth or the output of another model. We also randomize the position of the two videos to eliminate any potential bias that could arise from their placement. The survey is evenly distributed so that participants have an opportunity to view different video pairs for each emotion. The UI of the user study survey designed for perceptual evaluation is shown in Figure \ref{fig:user_study_desk}.

\end{document}